\pdfoutput=1

\documentclass[11pt]{article}

\usepackage[final]{acl}

\usepackage{times}
\usepackage{latexsym}

\usepackage[T1]{fontenc}

\usepackage[utf8]{inputenc}

\usepackage{microtype}

\usepackage{inconsolata}

\usepackage{graphicx}

\usepackage{listings}
\usepackage{url}            
\usepackage{booktabs}       
\usepackage{amsfonts}       
\usepackage{nicefrac}       
\usepackage{microtype}      
\usepackage{xcolor}         
\usepackage{multirow}
\usepackage{adjustbox}
\usepackage{amsmath}
\usepackage{pifont}
\usepackage{colortbl}
\usepackage{wrapfig}
\usepackage{makecell}
\usepackage{caption}
\usepackage{subcaption}
\usepackage{graphicx}
\usepackage{mathrsfs}
\usepackage{amssymb}
\usepackage{mathtools}
\usepackage{amsthm}
\usepackage{booktabs}
\usepackage{multirow}
\usepackage{listings}
\usepackage{xcolor}
\usepackage{algorithm}
\usepackage{algpseudocode}
\usepackage{graphicx}
\usepackage{float}
\usepackage{fancyvrb}
\usepackage{mdframed} 
\usepackage{verbatim}
\usepackage{tcolorbox}
\tcbuselibrary{listings}
\usepackage{tcolorbox}
\usepackage{color}
\tcbuselibrary{listings}
\usepackage{geometry}
\usepackage{enumitem}
\usepackage{calc}
\usepackage{tcolorbox}
\usepackage{textcomp} 
\tcbuselibrary{listings, breakable}
\usepackage{spverbatim}
\usepackage{array}
\usepackage{xspace}    
\usepackage{fancyhdr}
\usepackage{etoc}   
\usepackage{tabularx}
\usepackage{caption} 

\newcommand{\ladder}{\xspace\textsc{ladder}\xspace}

\newcommand\eg {{\it e.g., }}

\newcommand\etc {{etc.}}

\newcommand\ie {{\it i.e., }}

\definecolor{lightergray}{gray}{0.05}
\definecolor{boxgray}{rgb}{.94, .94, .94}
\newtcolorbox{mybox}[1][]{
  colback=white!95!gray, 
  colframe=gray!75!black, 
  fonttitle=\bfseries\large\ttfamily, 
  title=#1,
  coltitle=white, 
  colbacktitle=gray!75!black, 
  sharp corners,
  boxrule=1pt,
  left=4pt,
  right=4pt,
  top=4pt,
  bottom=4pt,
  breakable,
  width=\dimexpr\textwidth+60pt\relax, 
  enlarge left by=-0.5pt, 
  enlarge right by=-0.5pt, 
}
\definecolor{codebg}{rgb}{0.95,0.95,0.95}
\lstdefinestyle{mypython}{
    language=Python,
    frame=single,
    basicstyle=\footnotesize\ttfamily,
    keywordstyle=\color{blue},
    commentstyle=\color{gray},
    stringstyle=\color{red},
    showstringspaces=false,
    numbers=none  
}

\title{\ladder: Language-Driven Slice Discovery and Error Rectification in Vision Classifiers}

\author{
Shantanu Ghosh$^1$, Rayan Syed$^1$, Chenyu Wang$^1$, Vaibhav Choudhary$^1$, Binxu Li$^2$, Clare B Poynton$^3$, \\
\textbf{Shyam Visweswaran$^4$, Kayhan Batmanghelich$^1$} \\
$^1$Boston University, $^2$Stanford University, $^3$Boston University Medical Campus, $^4$University of Pittsburgh \\
\texttt{\{shawn24, rsyed, chyuwang, vchoudh , batman\}@bu.edu,  andy0207@stanford.edu,} \\
\texttt{Clare.Poynton@bmc.org}, \texttt{shv3@pitt.edu}
}

\begin{document}
\maketitle
\begin{abstract}
\label{sec:abstract}
Slice discovery refers to identifying systematic biases in the mistakes of pre-trained vision models. Current slice discovery methods in computer vision rely on converting input images into sets of attributes and then testing hypotheses about configurations of these pre-computed attributes associated with elevated error patterns. However, such methods face several limitations: 1) they are restricted by the predefined attribute bank; 2) they lack the \emph{common sense} reasoning and domain-specific knowledge often required for specialized fields \eg radiology; 3) at best, they can only identify biases in image attributes while overlooking those introduced during preprocessing or data preparation. We hypothesize that bias-inducing variables leave traces in the form of language (\eg logs), which can be captured as unstructured text. Thus, we introduce\ladder, which leverages the reasoning capabilities and latent domain knowledge of Large Language Models (LLMs) to generate hypotheses about these mistakes. Specifically, we project the internal activations of a pre-trained model into text using a retrieval approach and prompt the LLM to propose potential bias hypotheses. To detect biases from preprocessing pipelines, we convert the preprocessing data into text and prompt the LLM. Finally,\ladder generates pseudo-labels for each identified bias, thereby mitigating all biases without requiring expensive attribute annotations.
Rigorous evaluations on 3 natural and 3 medical imaging datasets, 200+ classifiers, and 4 LLMs with varied architectures and pretraining strategies -- demonstrate that\ladder consistently outperforms current methods. Code is available: \url{https://github.com/batmanlab/Ladder}.

\end{abstract}

\section{Introduction}
\label{sec:intro}

Error slices are data subsets on which vision classifiers systematically fail. Discovering such slices is critical for improving model robustness. Identifying such slices is challenging in vision classifers where biases are pervasive and can be traced through textual artifacts such as image captions, metadata, and medical imaging headers \eg DICOMs. However, their unstructured nature makes manual analysis impractical. Natural language, with its inherent flexibility, offers a powerful tool for capturing subtle biases beyond predefined attribute sets. LLMs, equipped with advanced reasoning capabilities and latent domain knowledge, excel at analyzing such free-form text to detect complex relationships and domain-specific biases. However, existing slice discovery methods often rely on predefined attribute banks or unsupervised clustering, both of which lack the reasoning ability to identify nuanced and domain-specific biases. This paper proposes\ladder, that leverages LLMs to systematically identify and mitigate error slices in vision classifiers by analyzing captions, metadata, and beyond -- without relying on fixed attribute sets or clustering methods.

\begin{figure}[t]
  \centering
  \includegraphics[width=0.4\textwidth]{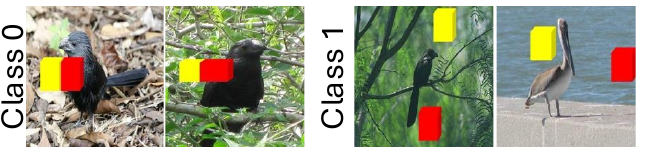}
  \caption{Synthetic dataset containing Class 0 images consistently with a yellow box to the left of a red box, while Class 1 images have boxes placed randomly. Captions encode the spatial bias, used by\ladder for slice discovery.
  }
  \label{fig:motivation}
  \vspace{-1.6em}
\end{figure}

\begin{figure*}[h]
\begin{center}
\includegraphics[width=\linewidth]{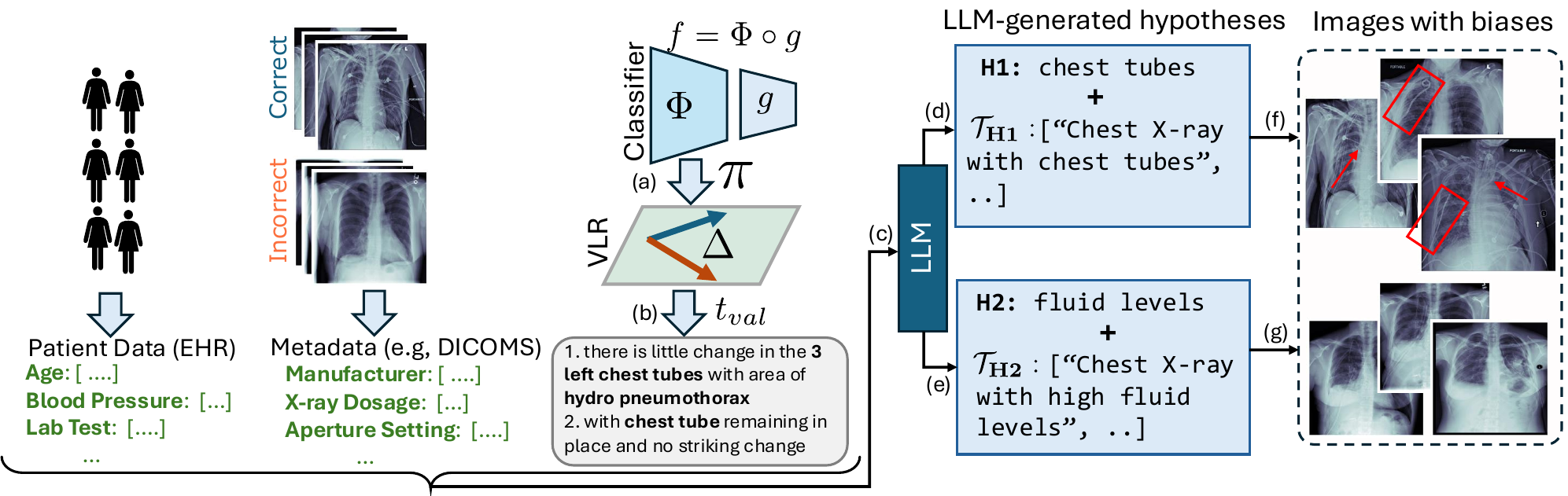}
\caption{Schematic of \ladder. \textbf{(a)}: Projection ($\pi$) of model representation ($\Phi$) to VLR space. \textbf{(b)}: Retrieval of \texttt{topK} sentences based on the image embeddings difference ($\Delta$) of correct and incorrect groups in VLR space. \textbf{(c)}: LLM is invoked with \texttt{topK} sentences/other metadata. \textbf{(d-e)}: LLM generated hypotheses (\{$\mathcal{H}, \mathcal{T}$\}). \textbf{(f-g)}:  Finding the clusters faithful to the hypotheses. In red, we highlight the chest tubes (ground truth bias for NIH) in this example.}
\label{fig:schematic}
\vspace{-1.2em}
\end{center}
\end{figure*}

Prior slice discovery methods \eg DrML~\citep{zhang2023diagnosing} use text encoders to mitigate biases in CLIP by closing the modality gap through cross-modal transfer, which limits their applicability to non-multimodal models. Plus, DrML relies on user-defined prompts with fixed attribute sets, introducing human bias into the mitigation process. Similarly, Facts~\citep{yenamandra2023facts} amplifies the spuriousness in the initial training stage by setting large weight decay, deviating from standard supervised learning practices. Methods like Domino~\citep{eyuboglu2022domino} and Facts discover slices by clustering samples with similar attributes within the vision-language representation (VLR) space. However, the slices often exhibit semantic inconsistencies -- attributes within slices lack coherence, leading to unreliable interpretations of model errors. PRIME~\citep{rezaei2023prime} relies on expensive tagging models, limited to detecting the presence/absence of a fixed set of attributes. All these methods lack the reasoning capabilities and domain knowledge required to capture complex error patterns, limiting their effectiveness in specialized tasks. Also, their dependence on pre-existing semantic labels (\eg visual tags) hinders the detection of biases in the metadata or domain-specific fields such as DICOM headers.


Prior mitigation methods~\cite{Sagawa2020, Liu2021, kirichenko2022last} rely on expensive and incomplete attributes. While they improve worst group accuracy (WGA), they amplify errors in other groups~\cite{li2023whac}. Although ~\citet{li2023whac} addresses errors across multiple biases, it assumes prior knowledge of the number and types of biases to design specific data augmentations. This reveals a critical gap: the need for an automated method to discover and mitigate multiple biases without prior knowledge/annotations.

This paper proposes\ladder with the following contributions:
\textbf{1. Using language for error slice discovery:}\ladder uses image captions/radiology reports to retrieve sentences indicative of model errors, utilizing the flexibility of natural language to capture deeper insights beyond the simple presence or absence of attributes, unlike tagging models.
\textbf{2. Using LLMs' reasoning capabilities and latent domain knowledge:} To identify biases,\ladder leverages LLMs' advanced reasoning to generate testable hypotheses from these sentences, unlike traditional methods.
For instance, in a synthetic dataset (Appendix~\ref{appendix_dataset_toy}), where Class 0 images consistently feature a yellow box to the left of a red box (Fig~\ref{fig:motivation}), the classifier exhibits poor performance on test data without this bias.\ladder correctly identifies this reliance on spatial positioning by analyzing textual descriptions through LLM (Fig~\ref{fig:toy}). Note, LLM in\ladder processes only text inputs without images (total cost of $\sim$\$28). In medical images,\ladder uses LLMs' domain knowledge to identify fine-grained biases, including disease subtypes and pathological patterns.
\textbf{3. Slice discovery from any off-the-shelf model:} It detects slices from any supervised model, regardless of architecture/pretraining, overcoming specific training requirements of Facts and DrML. 
\textbf{4. Detecting biases beyond captions:}\ladder uses LLM to analyze metadata, such as Electronic Health Records (EHR) or DICOM headers, discovering biases beyond captions.
\textbf{5. Mitigating multiple biases w/o any annotation:}\ladder mitigates biases by generating pseudo-labels for each hypothesis and fine-tuning the classifier's linear head through attribute rebalancing. By ensembling debiased model predictions,\ladder corrects multiple biases without requiring attribute annotations/prior knowledge of their number and type.
\textbf{Additionally,} we explore the use of instruction-tuning models (\eg LLaVA) in applicable domains to reduce\ladder's reliance on captions. Rigorous evaluations on 6 datasets with 200+ classifiers and 4 LLMs across architectures and pretraining strategies show that\ladder outperforms slice discovery and mitigation baselines.

\begin{figure*}[h]
\begin{center}
\includegraphics[width=0.9\textwidth]{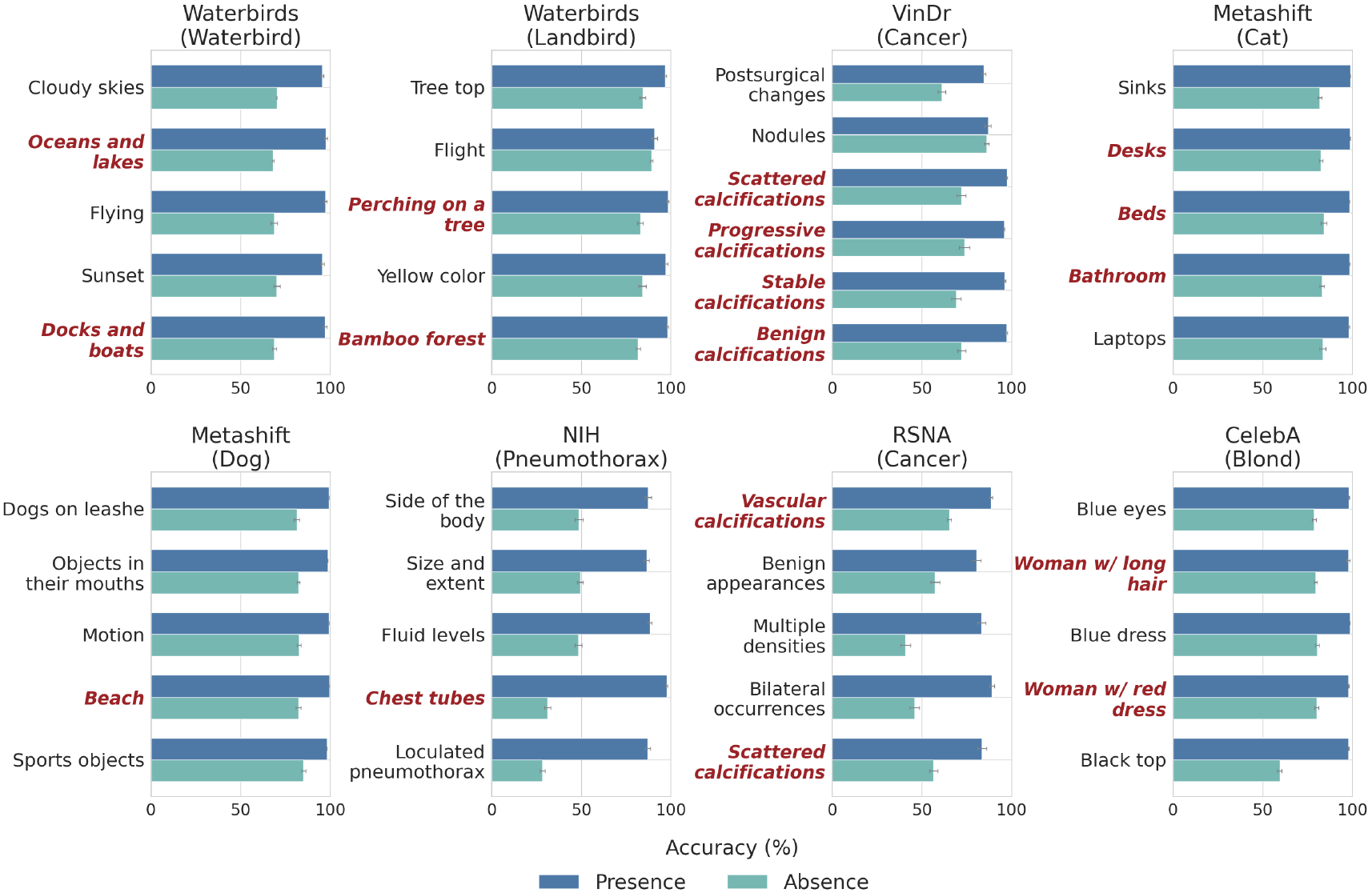}
\caption{
Bias identification by\ladder in RN Sup IN1k classifier. Each panel shows the classifier’s performance for a specific dataset (RSNA) and class label (Cancer) when biased attributes in the identified hypotheses are present/absent. Hypotheses indicative of ground truth biases (\eg water for waterbirds) are shown in red.}
\label{fig:qual-RN}
\end{center}
\vspace{-1em}
\end{figure*}

\section{Related Work}
\label{sec:related_work}
\textbf{Slice discovery.} Initial methods~\citep{d2022spotlight, sohoni2020no, kim2019multiaccuracy, singla2021understanding} on slice discovery utilize dimensionality reduction, lacking comprehensive evaluation. Recent methods \eg Domino~\citep{eyuboglu2022domino} projects data into VLR space, identifies slices via a mixture model, and captions them. Facts~\citep{yenamandra2023facts} amplifies spurious correlations in the initial training phase by increasing weight decay and discovering slices in VLR space. Both approaches compromise visual semantics, resulting in attribute inconsistencies within slices. DrML~\citep{zhang2023diagnosing} probes only CLIP-based classifiers using modality gap geometry and user-defined prompts, introducing potential human biases. Also, Facts and DrML are restricted to specific training setups, limiting generalizability to standard ERM classifiers. PRIME~\citep{rezaei2023prime} uses expensive tagging models to discover attributes for slice discovery. HiBug~\citep{chen2024hibug} prompts LLM to suggest biases for model errors without any textual context from the data. Thus, it results in superficial keyword-based attributes derived purely from general user prompts, lacking the deeper contextual grounding needed for bias detection. Recently, OpenBias~\citep{d2024openbias} detects biases in T2I models via LLM-driven keyword queries but is not designed for posthoc classifier error analysis. B2T~\citep{kim2024discovering} extracts keywords from captions. All these methods are limited by incomplete tags or keyword-based attributes and lack reasoning or latent \textit{domain knowledge}, essential in fields \eg radiology.
\textbf{Bias mitigation.} Mitigation methods \eg GroupDRO~\citep{Sagawa2020} optimizes for worst-performing groups, while JTT~\citep{Liu2021} reweights minority groups. DFR~\citep{kirichenko2022last} retrains the final layer using a balanced validation set. All of them require group annotations and focus on mitigating errors in the worst-performing group, amplifying errors in other subgroups.~\citet{li2023whac} mitigates multiple biases using an ensemble-based approach but relies on predefined bias types, which limits its adaptability to unknown biases.\ladder overcomes all these limitations. For discovery,\ladder incorporates the \textit{domain knowledge} of LLMs, reason about model errors, and generates hypotheses identifying biases from any pretrained model without external attributes, unlike existing methods. For mitigation,\ladder leverages pseudo-labels for each bias to finetune the classifier's last layer -- without any group annotations, predefined bias types, or human intervention. \textbf{Followup work.}
Our work is extended by~\cite{ghosh2024distributionally}, which adapts mitigation strategies for self-supervised learning on tabular data.

\section{Method}
\label{sec:method}
Assume the classifier $f=g \circ \Phi$ is trained using ERM to predict the labels $\mathcal{Y}$ from the images $\mathcal{X}$, where $\Phi$ and $g$ are the representation and classification head, respectively. $\{\Psi^I, \Psi^T\}$ denote the image and text encoders of the joint VLR space. For a set of images $\mathcal{X}_Y$ of a class $Y\in\mathcal{Y}$,\ladder identifies error slices where $f$ underperforms and mitigates it. Throughout the paper, $\langle\cdot,\cdot\rangle$ denotes the dot product to estimate the similarity between two representations. Fig.~\ref{fig:schematic} shows the schematic of\ladder. We do not rely on sample-specific paired annotations, human-generated prompts, or prior knowledge of bias types or their numbers. We utilize a text corpus $t_{val}$ from radiology reports or image captions from the validation dataset to discover and mitigate errors.
\textbf{Error slice.}
An error slice for a class $Y$ includes subsets $\mathcal{X}_Y$ where the model performs significantly worse than its overall performance on the entire class $Y$, formally defined as:
$
\mathbb{S}_{Y} = \{\mathcal{S}_{Y, \neg\texttt{attr}} \subseteq \mathcal{X}_Y|  e(\mathcal{S}_{Y, \neg\texttt{attr}}) \gg e(\mathcal{X}_Y), \exists \texttt{attr}\},
$
where $e(\cdot)$ is the error rate on the specific data subset and $\mathcal{S}_{Y, \neg\texttt{attr}}$ denotes the subset of $\mathcal{X}_Y$ without the attribute \texttt{attr}. Alternatively, $f$ is biased on the attribute \texttt{attr}, resulting in better performance on the subpopulation with \texttt{attr} \eg error rate in pneumothorax patients w/o chest tubes is higher than overall pneumothorax patients~\citep{docquier2012globalization}.

\begin{figure}[h]
  \centering
  \includegraphics[width=1.0\linewidth]{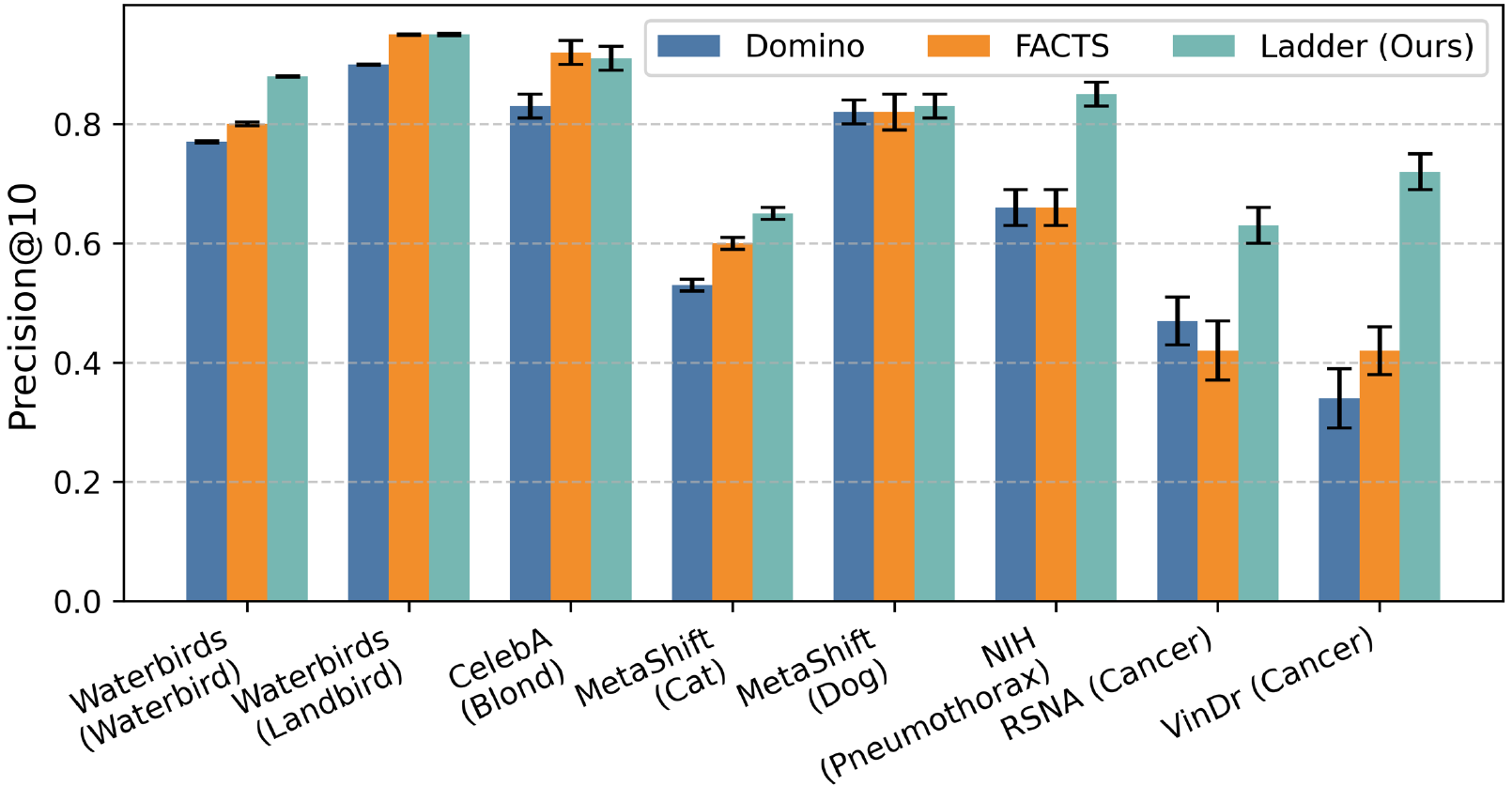}
  \caption{Precision@10 for CNN models ($f$) quantifying slice discovery.\ladder outperforms the baselines, especially for medical imaging datasets.}
  \label{fig:precision_10_main}
  \vspace{-1em}
\end{figure}

\subsection{Retrieving Sentences Indicative of Biases}
\label{sub:sent_retrieval}
First, for a particular class,\ladder retrieves the sentences that describe the visual attributes contributing to correct classifications but missing in misclassified ones, leading to model errors. Following~\citet{moayeri2023text}, it learns a projection function $\pi: \Phi \rightarrow \Psi^I$ (Appendix~\ref{appendix: projection}) to align the representation of the classifier, $\Phi$, with the image representation $\Psi^I$ of the VLR space. Then, for a class label $Y$, we estimate the difference in mean of the projected representations of the correct and misclassified samples as $\Delta^I = \mathbb{E}_{X, Y|f(X)=Y} [\pi(\Phi(X)] - \mathbb{E}_{X, Y|f(X)\neq Y} [\pi(\Phi(X)]$. Assuming the mean representations preserve semantics, this difference captures key attributes contributing to correct classifications but are poorly captured or misrepresented in misclassified ones. Denoting the text embedding of $t_{val}$ as $\Psi^T(t_{val})$, we retrieve the $\texttt{topK}$ sentences as: $
\texttt{topK} = \mathscr{R}\big(\langle\Delta^I, \Psi^T(t_{val})\rangle, t_{val}\big), 
$
where $\mathscr{R}$ is a retrieval function retrieving \texttt{topK} sentences from the text corpus having the highest similarity score with the mean difference of the projected image representations. Next, the LLM analyzes the sentences and constructs hypotheses to find error slices.

\subsection{Discovering Error Slices via LLM}
\label{sec:hypthosis}
\textbf{Generating hypothesis.}
To form the set of hypotheses,\ladder invokes an LLM with the \texttt{topK} sentences. Formally, $\{\mathcal{H}, \mathcal{T}\} = \texttt{LLM}\big(\texttt{topK})$, where $\mathcal{H}$ is a set of hypotheses with attributes that $f$ may be biased and $\mathcal{T}$ is a set of sentences to be used to test each hypothesis. $f$ underperforms on the subpopulation without the attributes in $\mathcal{H}$. Each hypothesis $H\in\mathcal{H}$ is paired with $\mathcal{T}_H\in\mathcal{T}$, a set of sentences that provide diverse contextual descriptions of the hypothesis-specific attribute as it appears in various images. Representations of images with the attribute specified in $H$, are highly similar to the mean text embedding of $\mathcal{T}_H$. Refer to Appendix~\ref{appendix: prompts} for the prompt used by LLM to generate the hypothesis.
\textbf{Identifying error slices.}
For each hypothesis $H \in \mathcal{H}$, we first compute the mean embedding of the set of sentences $\mathcal{T}_H$ as $\Psi^T(\mathcal{T}_H) = \frac{1}{|\mathcal{T}_H|}\sum_{t\in\mathcal{T}_H}\Psi^T(t)$. For an image $X\in\mathcal{X}_Y$, we obtain the projected representation $\pi(\Phi(X))$ in VLR space and compute the similarity score,
$
s_H(X) = \langle\pi(\Phi(X)), \Psi^T(\mathcal{T}_H)\rangle
$
Finally, for a class label $Y$, we retrieve images with similarity scores below a threshold $\tau$ as $\mathcal{S}_{Y, \neg H} = \{X \in \mathcal{X}_Y | s_H(X) < \tau\}$. 
The hypothesis $H$ fails in these images as they lack the attribute specified in the $H$.
The subset $\mathcal{S}_{Y, \neg H}$ may be a potential error slice if the error $e(\mathcal{S}_{Y, \neg H})$ is greater than $\mathcal{X}_Y$. Formally, $\hat{\mathbb{S}}_Y$, the predicted slice for a class $Y$ is:
$\hat{\mathbb{S}}_{Y} = \{\mathcal{S}_{Y, \neg H} \subseteq \mathcal{X}_Y|  e(\mathcal{S}_{Y, \neg H}) \gg e(\mathcal{X}_Y), \exists H\in \mathcal{H} \}$


\subsection{Mitigate Multi-bias w/o Annotation}
\label{sub: bias_mitigation}
For the attributes linked to a hypothesis,\ladder treats \( s_H \) as a logit and converts it to a probability. If the probability exceeds a threshold (0.5 in all experiments),\ladder assigns a pseudo-label 1 to the attribute and 0 otherwise. Thus, it generates pseudo-labels for all relevant attributes, enabling error mitigation without annotations. To do so,\ladder adopts an ensemble-based strategy.
Following DFR, we create a balanced dataset from a held-out validation set, for each pseudo-labeled attribute per hypothesis. We then fine-tune the classification head $g$ using this balanced dataset, producing a debiased model per hypothesis.
During inference, we again compute the similarity score \( s_H \) for all hypotheses and select the classifier head \( g_{H^*} \) associated with the hypothesis having maximum similarity:
$H^* = \arg\max_{H \in \mathcal{H}} s_H(X)$.

\section{Experiments}
\label{sec:experiments}
We perform experiments to answer the research questions:
\textbf{RQ1.} How does\ladder perform in discovering error slices compared to baselines?
\textbf{RQ2.} How does\ladder leverage reasoning and latent domain knowledge of LLMs for slice discovery?
\textbf{RQ3.} How does\ladder discover biased attributes with different architectures and pre-training methods?
\textbf{RQ4.} How does\ladder mitigate biases using the discovered attributes?
\textbf{RQ5.} Can\ladder operate w/o captions?
\textbf{RQ6.} Can\ladder detect biases beyond captions/reports?

\textbf{Datasets.} We evaluate \ladder on 6 datasets (Appendix\ref{appendix_dataset} for details):
1) \textbf{Waterbirds}~\cite{Wah2011}: bird classification where background correlates with bird type.
2) \textbf{CelebA}~\cite{liu2018large}: blond hair classification with gender as a spurious feature.
3) \textbf{MetaShift}: cat vs.\ dog classification with background correlation.
4) \textbf{NIH Chest-X-ray (CXR)}~\cite{wang2017chestx}: pneumothorax detection with chest tubes as a shortcut~\cite{docquier2012globalization}.
5) \textbf{RSNA-Mammo} and 6) \textbf{VinDr-Mammo}~\cite{nguyen2023vindr}: breast cancer and abnormality detection from mammograms, with calcifications as a shortcut~\cite{wen2024breast}.

\textbf{Experimental details.} \label{sub:experimental_details} For natural images and CXRs, we use an ImageNet1k (IN1k)-initialized ResNet50 (RN Sup IN1k) as the model $f$ that\ladder aims to probe, trained with a standard supervised loss. For mammograms, we use EfficientNet-B5 (EN-B5) as $f$. For the text corpus ($t_{val}$), we use BLIP-captioner~\citep{li2022blip}, radiology reports from MIMIC-CXR~\citep{johnson2019mimic} and the radiology texts from Mammo-FActOR~\citep{ghosh2024mammo} for natural images, CXRs and mammograms, respectively. For VLR space ($\{\Psi^I, \Psi^T\}$), we use CLIP~\cite{radford2021learning}, CXR-CLIP~\citep{you2023cxr}, and Mammo-CLIP~\cite{ghosh2024mammo} for natural images, CXR and mammograms, respectively. We use 200 and 100 sentences as \texttt{topK} for natural and medical images (CXR and mammo). We use GPT-4o~\citep{wu2024gpt} as the LLM. Error slices are defined as subsets where the error rate exceeds the overall class error by at least 10\%. Refer to Appendix~\ref{appendix_experiments} for further experimental details. All reported results are obtained from experiments conducted over 3 random seeds.

\textbf{Baselines.} For slice discovery, we compare\ladder with Domino and Facts (Appendix~\ref{appendix_slice_discovery}). For mitigation, we compare with the baselines, including ERM~\citep{Vapnik1999}, GroupDRO~\citep{Sagawa2020}, JTT~\citep{Liu2021}, DFR~\citep{Imazilov2022}, CVaRDRO~\cite{Duchi2018} and LfF~\cite{Nam2020} (Appendix~\ref{appendix_rr_mitigation}). 

\textbf{Evaluation metrics.} We use \texttt{Precision@10} (Appendix~\ref{appendix: precision@k})~\citep{eyuboglu2022domino} to evaluate the slice discovery methods and the CLIP score~\citep{kim2024discovering} to quantify the effect of biased attributes. For mitigation, we report Worst Group Accuracy (WGA) for mitigation for natural images. We report mean AUROC and WGA for medical images, where WGA refers to model performance on pneumothorax patients w/o chest tubes (NIH) and cancer or abnormal patients w/o calcifications (RSNA \& VinDr-Mammo).

\section{Results}
\label{sec:results}
\textbf{RQ1: Comparison of\ladder with slice discovery baselines.}
Following~\cite{eyuboglu2022domino, yenamandra2023facts}, Fig.~\ref{fig:precision_10_main} compares the \texttt{Precision@10} of different slice discovery methods for CNN models (EN-B5 for mammograms \& RN Sup IN1k for others). For medical images,\ladder achieves a substantial \textbf{50\%} improvement over the baselines. Refer to Fig.~\ref{fig:domino_compare_ladder_wga} in Appendix~\ref{appendix: domino-facts-ladder} for WGA evaluation using the slices discovered from Domino, Facts, and~\ladder with our ensemble-based mitigation strategy. In all the experiments,\ladder outperforms the baselines.
Facts and Domino cluster the images by projecting them directly into VLR space, often leading to incoherent slices. In contrast,\ladder first projects the model's representation into the VLR space, preserving the nuanced semantics of the classifier features. Instead of relying solely on unsupervised clustering, it leverages the reasoning capabilities of LLMs and signals from the captions/radiology reports to identify the coherent-biased attributes within the discovered slices. Next, we assign pseudo-labels to the attributes using similarity scores ($s_{H}(X)$). The coherent slices produced by\ladder ensure that the pseudo-labeling process is more accurate than the baselines leading to superior bias mitigation performance (Appendix~\ref{appendix: domino-facts-ladder}).

\begin{figure}[t]
\begin{center}
\includegraphics[width=\linewidth]{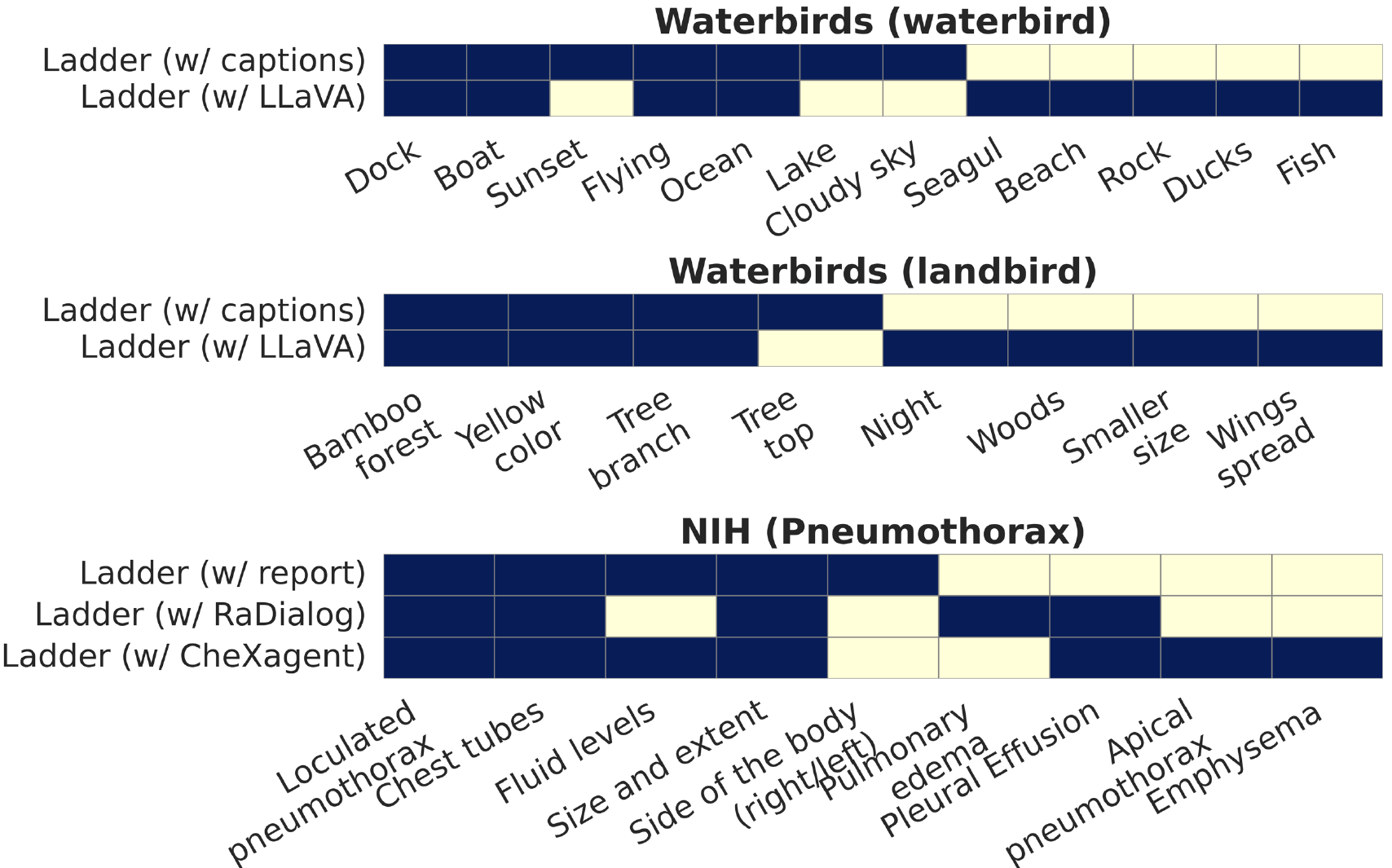}
\caption{Biased attributes detected by LADDER w/
captions and w/ instruction-tuned models (w/o captions).
Bright/light colors show presence/absence of attributes}
\label{fig:llava-heatmap}
\end{center}
\vspace{-1.1em}
\end{figure}

\begin{figure}[h]
  \centering
  \includegraphics[width=\linewidth]{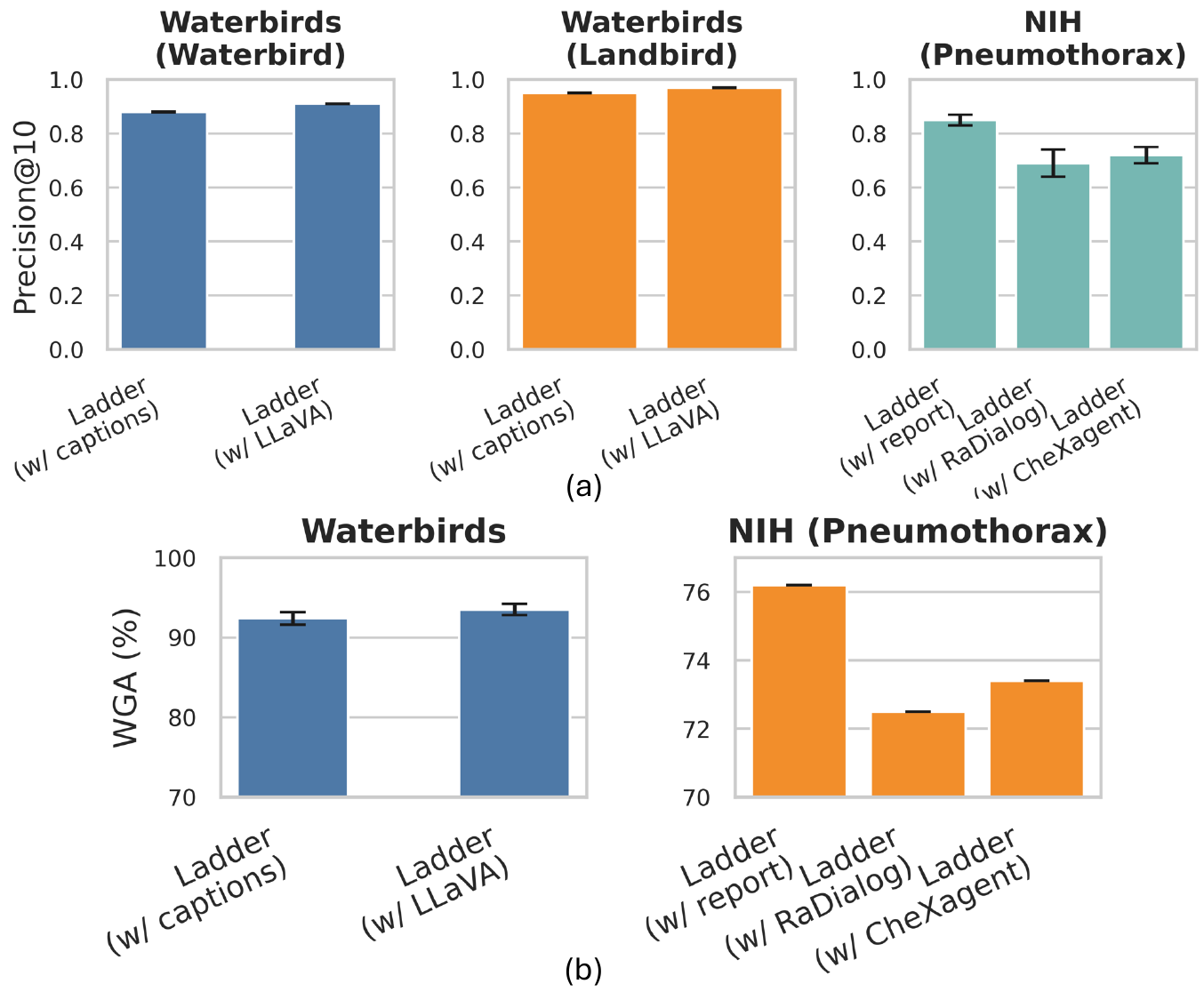}
  \caption{(a) Precision@10 for slice discovery and (b) WGA for bias mitigation using\ladder w/ captions vs. instruction-tuned models.}
  \label{fig:llava-precision-mit}
  \vspace{-1.2em}
\end{figure}

\begin{figure}[h]
  \centering
  \includegraphics[width=\linewidth]{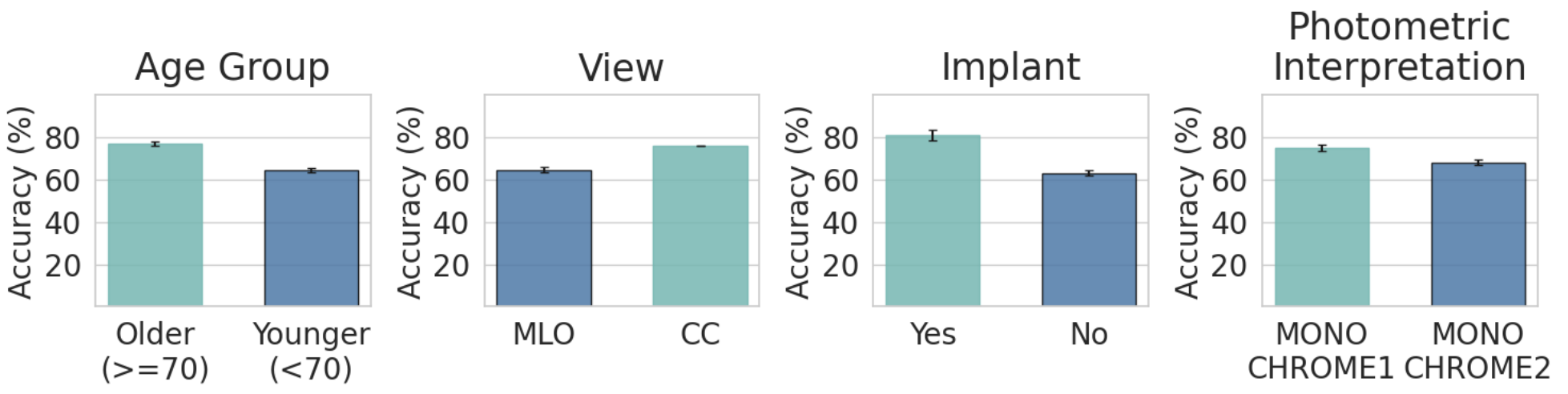}
  \caption{\ladder detects biases beyond reports,  identifying biases from metadata (age, view and implant) and DICOM headers (Photometric interpretation).}
  \label{fig:metadata-bias}
\vspace{-0.5em}
\end{figure}

\begin{figure}[t]
\begin{center}
\includegraphics[width=0.95\linewidth]{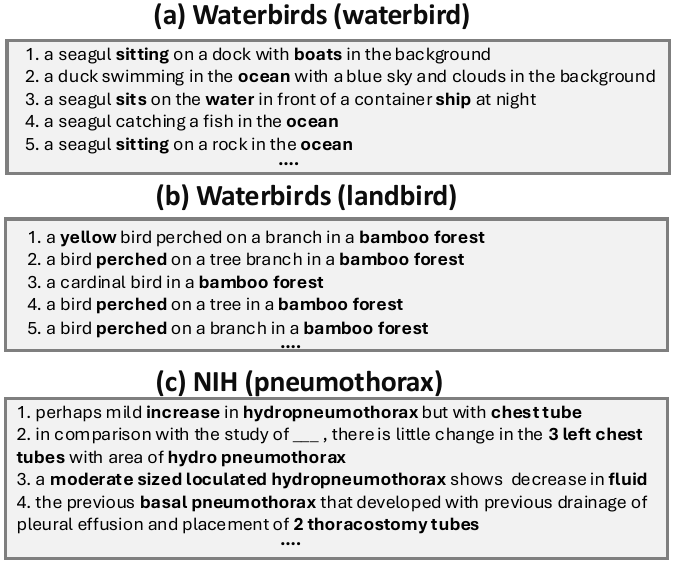}
\caption{Sentences retrieved by\ladder in Sec.~\ref{sub:sent_retrieval} encoding model biases (in bold) for LLM to analyze. Each panel denotes a class label of a specific dataset.
} 
\label{fig:sent}
\end{center}
\vspace{-2.em}
\end{figure}

\noindent\textbf{RQ2. Leveraging LLM's reasoning and domain knowledge for bias discovery.} 
\label{results:nih_waterbirds} Fig.~\ref{fig:sent} displays the sentences retrieved by\ladder indicating the different model biases. Fig.~\ref{fig:qual-RN} shows the biased attributes discovered by\ladder. The presence of these attributes correlates with $f$’s performance, while their absence results in error slices where $f$'s performance drops. Recall,\ladder uses LLM to generate hypotheses from the sentences, indicative of biases. The similarity score ($s_H(X)$) tests these hypotheses to validate if the absence of specific attributes linked to each hypothesis results in a drop in $f$'s performance. For \eg waterbirds \texttt{flying} vs. \texttt{not flying} achieve 97.3\% vs. 68.6\% accuracy. In NIH, pneumothorax patients with and without \texttt{chest tubes} achieve an accuracy of $\sim$98\%, compared to 31\%. 
For all tasks,\ladder effectively detects ground truth biases. In the Waterbirds dataset,\ladder identifies diverse water-related biases such as \texttt{boat} and \texttt{lake}. Also, Fig.\ref{fig:qual-RN} reports that\ladder identifies domain-specific biases (\eg \texttt{chest tubes}, \texttt{loculated pneumothorax} for NIH; subtypes of \texttt{calcifications} for RSNA \& VinDr Mammo), capturing a more granular characterization of biases. Unlike the keyword extraction or tagging models, which struggle with missing or insufficient attributes,\ladder leverages LLM-driven latent medical knowledge to generate comprehensive hypotheses. Such fine-grained detection of contextual biases, including subtypes, allows\ladder to for the detection of patterns that would be difficult to detect without domain expertise. Refer to Appendix~\ref{appendix: extended_qual},~\ref{appendix: closest_attr} and~\ref{clip_score} for detailed qualitative results, the hypotheses closest to the ground truth biases, and the influence of biased attributes via CLIP score, respectively.

\begin{table}[h]
\scriptsize
\centering
\caption{Impact of captioners on\ladder's performance for RN Sup IN1k classifier. Though GPT-4o is expensive, its quality is better than others.}
\begin{tabular}{l@{\hskip 6pt}c@{\hskip 6pt}c@{\hskip 6pt}c@{\hskip 6pt}c}
\toprule
 & \multicolumn{2}{c}{Waterbirds} & \multicolumn{2}{c}{CelebA} \\
\cmidrule(lr){2-3} \cmidrule(lr){4-5}
Method & Mean Acc & WGA & Mean Acc & WGA \\
\midrule
BLIP~\citep{li2022blip} & 93.1 & 91.4 & 89.8 & 88.9 \\
BLIP2~\citep{li2023blip} & 93.3 & 91.6 & 89.8 & 89.2 \\
ClipCap~\citep{mokady2021clipcap} & 93.7 & 91.8 & 88.3 & 87.4 \\
GPT-4o~\citep{wu2024gpt} & \textbf{94.2} & \textbf{93.1} & \textbf{91.4} & \textbf{90.3} \\
\bottomrule
\end{tabular}
\vspace{-1.2em}
\label{tab:captioning_ablation}
\end{table}

\begin{figure*}[h]
\begin{center}
\includegraphics[width=\linewidth]{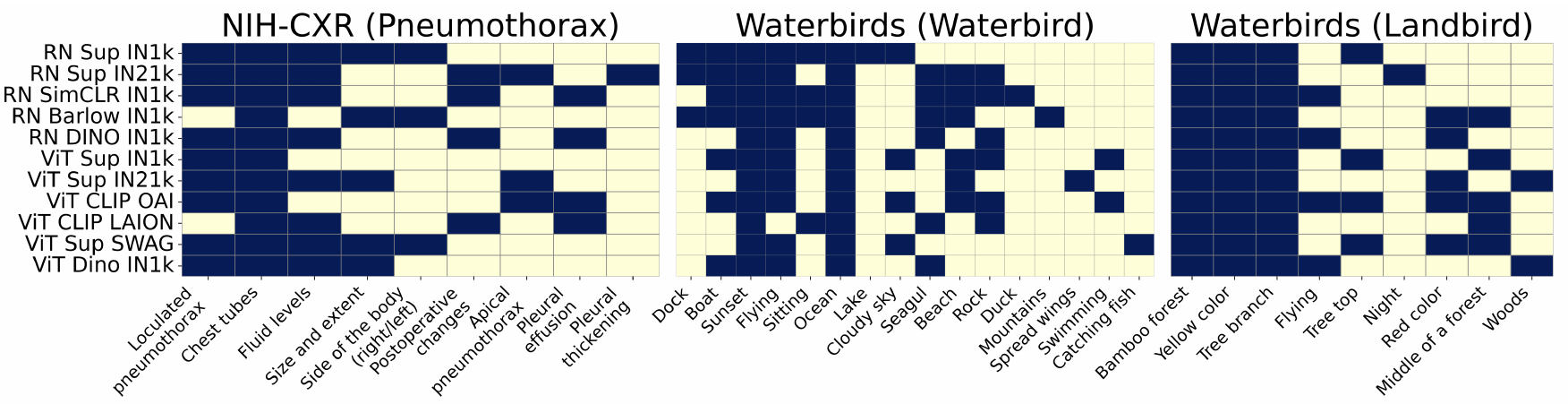}
\caption{Biased attributes discovered by \ladder show consistent biases across architectures and pretraining. Several attributes (\eg ocean, lake, beach \etc) represent the same visual concepts (water bodies) denoting the groundtruth bias. Bright and light colors indicate attribute presence and absence, respectively.} 
\label{fig:hyp_main_all}
\end{center}
\vspace{-1em}
\end{figure*}

\noindent\textbf{RQ3: Biased attributes discovery across architectures/pre-training methods.} In this setup, we extract biases using\ladder on a range of model architectures (both ResNet50 and ViT), initializing $f$ (the model to be probed) with diverse pretraining methods, including SimCLR~\citep{chen2020simple}, Barlow Twins~\citep{zbontar2021barlow}, DINO~\citep{caron2021emerging}, and CLIP~\citep{radford2021learning}. These methods are pretrained on datasets \eg ImageNet-1K (IN1k)~\citep{deng2009imagenet}, ImageNet-21K (IN21k)~\citep{ridnik2021imagenet}, SWAG~\citep{singh2022revisiting}, LAION-2B~\citep{schuhmann2022laion}, and OpenAI-CLIP (OAI)~\citep{radford2021learning}.~\citet{yang2023change} shows that every ERM-trained classifier ($f$) exhibits low WGA irrespective of architecture/pretraining due to consistently learning similar biases. Figure \ref{fig:hyp_main_all} shows that\ladder, leveraging LLM-driven reasoning and \textit{domain knowledge}, consistently identifies similar biases across different architectures, pretraining methods, and datasets. In the NIH dataset,\ladder identifies mostly key attributes such as \texttt{chest tubes}, \texttt{fluid levels} \etc ~Also, in the Waterbirds dataset,\ladder detects attributes \eg \texttt{ocean} and \texttt{bamboo forest} consistently, showing the correlation of the spurious backgrounds with class labels and the ground truth biases. Appendix~\ref{appendix: slices_arch_pretraining} lists more results. 

\noindent\textbf{RQ4: Mitigating biases using\ladder.}
Tab.~\ref{tab:mitigation_RN50} shows that\ladder outperforms other bias mitigation baselines in estimating WGA, without requiring the expensive ground truth shortcut attributes, for both training and validation datasets across CNN models (EN-B5 for Mammograms and RN Sup IN1k for the rest).\ladder achieves a WGA of 91.4\%, 76.4\% and 82.5\% -- a 3.6\%, 7.3\% and 21.1\% improvement ($\uparrow$) over DFR
in the Waterbirds, RSNA, and VinDr datasets, respectively.
For NIH,\ladder outperforms JTT and DFR by 8.2\% and 7.4\%, respectively. Appendix~\ref{app:mitigation_extended} illustrates further analysis with an additional 9 baselines. Fig.~\ref{fig:wga_all} shows\ladder's consistent performance gain across various architectures and pre-training methods. Tab.~\ref{tab:whacamole} in Appendix~\ref{app: whac} shows that\ladder outperforms~\citet{li2023whac} on multi-shortcut benchmark UrbanCars.
Leveraging LLMs' advanced reasoning,\ladder accurately derives pseudo labels for the biased attributes from hypotheses to identify true model biases.\ladder then applies targeted bias mitigation by fine-tuning the last layer, resulting in a systematic debiased model per hypothesis. This efficient strategy effectively enhances model performance across the biases, modalities, and architectures.

\begin{table}[h!]
\scriptsize
\centering
\caption{Error mitigation results (WGA) for EN-B5 for mammograms and RN Sup IN1k for the rest. We bold-face and underline the best and second-best results. We compare with 9 additional baselines in ~\ref{app:mitigation_extended}.}
\label{tab:mitigation_RN50}
\setlength{\tabcolsep}{3pt}
\begin{adjustbox}{max width=\textwidth}
\begin{tabular}{lcccccccccc}
\toprule
\multicolumn{1}{c}{\textbf{Method}} & 
\multicolumn{1}{c}{\textbf{Waterbirds}} 
& \multicolumn{1}{c}{\textbf{CelebA}} 
& \multicolumn{1}{c}{\textbf{NIH}} 
& \multicolumn{1}{c}{\textbf{RSNA}} 
& \multicolumn{1}{c}{\textbf{VinDr}}\\
\midrule
ERM
& 69.1$_{\pm1.2}$ 
& 62.2$_{\pm1.5}$ 
& 60.3$_{\pm0.0}$ 
& 69.8$_{\pm0.0}$
& 45.6$_{\pm0.0}$ \\

JTT 
&  84.5$_{\pm0.3}$ 
&  87.2$_{\pm7.5}$  
&  70.4$_{\pm0.0}$  
&  68.5$_{\pm0.0}$ 
&  66.1$_{\pm0.0}$ \\

GroupDRO 
&  87.1$_{\pm1.3}$ 
& \underline{88.1}$_{\pm0.7}$    
& 71.1$_{\pm0.0}$
& \underline{72.3}$_{\pm0.0}$ 
& 67.1$_{\pm0.0}$ \\

CVaRDRO 
& 85.4$_{\pm2.3}$ 
& 83.1$_{\pm1.5}$  
& \underline{71.3}$_{\pm0.0}$ 
& 71.7$_{\pm0.0}$ 
& 67.1$_{\pm0.0}$ \\

LfF 
& 75.2$_{\pm0.7}$ 
& 63.0$_{\pm4.4}$  
& 61.6$_{\pm0.0}$ 
& 66.4$_{\pm0.0}$ 
& 64.5$_{\pm0.0}$ \\

DFR 
& \underline{88.2}$_{\pm0.3}$ 
& 87.1$_{\pm1.1}$  
& {70.5}$_{\pm0.0}$   
& 71.2$_{\pm0.0}$ 
& \underline{68.1}$_{\pm0.0}$ \\
\midrule

\cellcolor{lightgray}\ladder
&\cellcolor{lightgray} \textbf{91.4}$_{\pm0.8}$ 
& \cellcolor{lightgray} \textbf{88.9}$_{\pm0.4}$ 
& \cellcolor{lightgray}\textbf{76.2}$_{\pm0.0}$  
& \cellcolor{lightgray}\textbf{76.4$_{\pm0.0}$}
& \cellcolor{lightgray}\textbf{82.5$_{\pm0.0}$}\\
\bottomrule
\end{tabular}
\end{adjustbox}
\end{table}

\noindent\textbf{RQ5: Relaxing the dependency on captions.}
To reduce\ladder’s reliance on captions/reports, we leverage instruction-tuned models to generate textual descriptions for the correctly classified samples. Specifically, we use LLaVA-1.5 7B~\cite{liu2024visual} for natural images and RaDialog \cite{pellegrini2023radialog} and cheXagent~\cite{chen2024chexagent} for CXRs to probe RN Sup IN1k classifier. Refer to Appendix~\ref{appendix: llava_med_prompt} for the utilized prompts. \ladder’s LLM pipeline utilizes these generated descriptions to identify biased attributes. Recall we aim to detect biases consistently present in correctly classified instances. Figure~\ref{fig:llava-heatmap} compares the biases identified using\ladder’s retrieval pipeline (captions/reports) vs. those detected via instruction-tuned models. Figure~\ref{fig:llava-precision-mit}(a) compares Precision@10 for \ladder under both settings, while Figure~\ref{fig:llava-precision-mit}(b) evaluates the WGA metric, evaluating the bias discovery and mitigation quantitatively, respectively. For natural images,\ladder with instruction-tuned models perform comparably to the standard pipeline using captions. For CXRs, the retrieval-based approach utilizing actual reports outperforms methods using cheXagent and RaDialog, highlighting the importance of domain-specific reports in medical imaging. Thus, using models \eg LLaVA can eliminate\ladder's need for captions. However, this approach is challenging for 2D mammograms and dermatology imaging~\cite{alzubaidi2021novel} \etc where robust instruction-tuned models are lacking. In such cases,\ladder's retrieval pipeline remains highly adaptable and shows broad applicability. Thus, a trade-off emerges: models can either leverage explicit radiology reports for bias identification or develop robust VLRs to reduce dependence on reports.

\begin{figure}[h]
\begin{center}
\includegraphics[width=\linewidth]{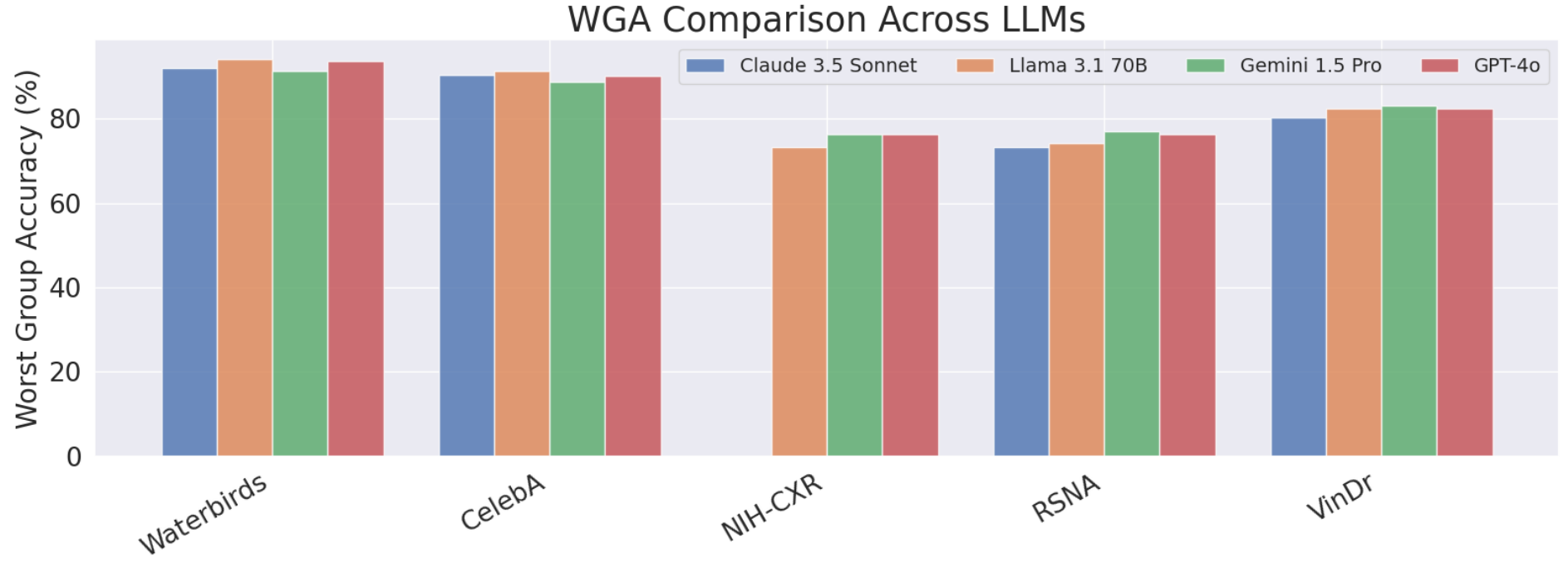}
\caption{WGA comparison across different LLMs for bias mitigation by\ladder with RN Sup IN1k for natural images and CXRs, and EN-B5 for mammograms. GPT-4o and Gemini excel in medical imaging tasks.} 
\label{fig:LLM_WGA}
\end{center}
\vspace{-1em}
\end{figure}

\noindent\textbf{RQ6: Detecting biases beyond captions/reports.}
While prior work~\cite{boyd2023potential} highlights biases in EHR and medical imaging headers (\eg DICOMs),\ladder extends bias detection beyond captions. 
We use metadata from the RSNA-Mammo dataset, which includes metadata \eg BIRADS(0-2), age, implant, view (CC or MLO), laterality (left or right breast), machine\_id, and site\_id. Also, the DICOM headers provide attributes \eg photometric interpretation, VOI LUT, and pixel intensity relationships. We probe the same EN-B5 classifier to find attributes consistently present in correctly classified samples, whose absence results in a performance drop. By listing each sample’s metadata to a Python dictionary (refer to Appendix~\ref{appendix: metadatabias}) and using\ladder’s LLM pipeline (Sec.~\ref{sec:hypthosis}), we generate hypotheses about the biased attributes; we then validate their impact on the classifier's performance based on the presence/absence of these attributes, with their ground truth values from the metadata. Figure~\ref{fig:metadata-bias} shows that\ladder detects an age bias (a 19.5\% accuracy gap for patients aged 70+ vs. the rest) and a 10\% gap to different photometric interpretations (Monochrome 1 vs. Monochrome 2). This finding aligns with existing evidence of age bias in oncology~\cite{tasci2022bias}. 
Existing methods lack LLM-based reasoning, limiting them to fixed attributes or clustering, while \ladder uses LLMs to reason across metadata for comprehensive analysis.

\section{Ablations and Additional Results}
Table~\ref{tab:captioning_ablation} compares\ladder's performance across different captioning methods, while Fig.\ref{fig:LLM_WGA} presents the WGA of\ladder for various LLMs. Due to space constraints, we provide detailed analyses in Appendices~\ref{WGA_captioning} and~\ref{WGA_LLM}. Additionally, Appendices~\ref{App:slice_LLM},~\ref{sec:cost_LLM} and~\ref{sec:ablation_cxr} include ablation studies on slices discovered using different LLMs, their computational costs, and the impact of different VLRs on\ladder. Appendix~\ref{sec: imagenet} demonstrates \ladder's ability to identify biases in the ImageNet dataset (multiclass classification), while Appendix~\ref{app: zs_boost} shows how these identified attributes improve CLIP's zero-shot accuracy.

\section{Conclusion}
\label{sec:conclusion}
We introduce\ladder, a novel LLM-driven method for error slice discovery and bias mitigation for vision classifiers. Unlike prior methods that rely on predefined attributes or unsupervised clustering,\ladder leverages LLM's reasoning to detect coherent error slices without requiring explicit annotations from any off-the-shelf pretrained classifier. Next, it mitigates multiple biases through pseudo-label generation and attribute rebalancing. Extensive evaluations on 6 datasets show \ladder’s effectiveness, outperforming existing baselines.




\section{Acknowledgments}
\label{sec:acknowledgment}
This work was partially supported by the Pennsylvania Department of Health, NIH Award Number 1R01HL141813-01, and funding from the Hariri
Institute for Computing, Boston University. We are grateful for the computational
resources from Pittsburgh Super Computing grant number TG-ASC170024.

\section*{Limitations}
\label{sec:limitations}
While\ladder demonstrates superior performance in bias discovery and mitigation, we outline the limitations of our work and potential areas for improvement:
\textbf{1. Dependence on captions for bias discovery:}\ladder primarily relies on captions to identify biases, which may not be suitable for domains with sparse or limited textual descriptions. While we introduce a workaround using instruction-tuned models \eg LLaVA for specific applications, future research will explore reducing language dependence across broader domains.
\textbf{2. Potential bias in pretrained models:}\ladder utilizes pretrained models such as CLIP and LLMs, which inherently reflect biases present in their training data. This dependency may influence the bias discovery process and potentially undermine fairness objectives. Addressing and mitigating these inherent biases in foundational models is an important direction for future research.
\textbf{3. Lack of human oversight in bias discovery:} To prevent the introduction of additional bias,\ladder automates the discovery phase without human intervention. Instead, domain experts (\eg clinicians) validate the identified biases prior to mitigation. While this strategy minimizes human-induced bias during discovery, it introduces subjectivity in the validation phase. Enhancing and standardizing this validation process remains a key focus for future work.

\section*{Ethical Considerations}
\label{sec:ethics}

We strongly adhere to ethical standards in the handling of medical data, the use of language models, and the implementation of machine learning methods. We provide the following details:
\textbf{1. Medical datasets:} All medical datasets used in this study, including MIMIC-CXR, RSNA-Mammo, and VinDr-Mammo, are anonymized and publicly available. We strictly follow the respective data-use agreements and ethical guidelines associated with each dataset.
\textbf{2. Language models for medical tasks:} The large language models (LLMs) employed for medical applications adhere to the guidelines established for MIMIC\footnote{\url{https://physionet.org/news/post/gpt-responsible-use}}. Specifically, we use GPT-4o~\citep{wu2024gpt} via Azure OpenAI service as LLM for NIH in the main experiments. For ablations, we use Google's Gemini via Vertex AI. For LLaMA, we set up the model on a local machine.  No information from NIH datasets was processed using language models not covered by these guidelines, such as Claude.
\textbf{3. Classifier models and codebase:} All classifiers used in this research are standard architectures and publicly available models, ensuring reproducibility and transparency. We list them in detail in Appendix~\ref{appendix_experiments}.
\textbf{4. Vision-Language representations (VLRs):} All VLRs utilized in this study are publicly available, and we list the corresponding resources in Appendix~\ref{appendix_experiments}. We adhere strictly to the license terms specified by the creators of these resources.

\section*{Broader Impact}
\label{sec:broader_impacr}
The development and deployment of\ladder have potential implications for AI applications in medical and general computer vision tasks. We outline the broader impacts as follows:
\textbf{1. Medical applications and patient outcomes:}\ladder can improve the robustness and interpretability of vision models in medical imaging. By identifying and mitigating biases, it can lead to more reliable diagnostic tools, ultimately enhancing patient care and reducing diagnostic disparities.
\textbf{2. Bias detection and fairness:}\ladder offers a generalizable approach to uncovering and addressing systematic biases across datasets. This can contribute to the development of fairer AI models, particularly in domains prone to dataset biases, such as healthcare and social applications.
\textbf{3. Continuous auditing and bias mitigation:}\ladder can act as an auditor for any pretrained network in a continuous manner. By running it on a dataset, it can identify and mitigate biases using language. Whenever a bias can be traced in language,\ladder can detect it with its superior reasoning capabilities and domain knowledge.


\bibliography{main}

\appendix

\section{Appendix}
\label{sec:appendix}


\subsection{ Extended details on datasets}
\label{appendix_dataset}
\subsubsection*{Waterbirds}
The \textbf{Waterbirds} dataset~\citep{Wah2011} is frequently employed in studies addressing spurious correlations. This binary classification dataset overlaps images from the Caltech-UCSD Birds-200-2011 (CUB) dataset with backgrounds sourced from the Places dataset \citep{Zhou2017}. The primary task involves determining whether a bird depicted in an image is a landbird or a waterbird, with the background (water or land) as the spurious attribute. For consistency and comparability, we adhere to the train/validation/test splits utilized in prior research \citep{Idrissi2022}.

\subsubsection*{CelebA}
The \textbf{CelebA} dataset~\citep{Liu2015} comprises over 200,000 images of celebrity faces. In the context of spurious correlations research, this dataset is typically used for the binary classification task of predicting hair color (blond vs. non-blond), with gender serving as the spurious correlation. In alignment with previous studies \citep{Idrissi2022}, we use the standard dataset splits. The CelebA dataset is available under the Creative Commons Attribution 4.0 International license.

\subsubsection*{MetaShift}
The \textbf{MetaShift} dataset~\citep{Liang2022} offers a flexible platform for generating image datasets based on the Visual Genome project \citep{Krishna2016}. Our experiments utilize the pre-processed \textit{Cat vs. Dog} dataset, designed to differentiate between cats and dogs. The dataset features the image background as a spurious attribute, with cats typically appearing indoors and dogs outdoors. We use the "unmixed" version of this dataset, as provided by the authors' codebase.

\subsubsection*{NIH chestXrays}
The \textbf{NIH} ChestX-ray dataset~\citep{wang2017chestx}, also known as ChestX-ray14, is a large dataset of chest radiographs (X-rays) provided by the National Institutes of Health (NIH).
The dataset comprises 112,120 frontal-view X-ray images of 30,805 unique patients. Each image is associated with one or more of the 14 labeled thoracic diseases, which include atelectasis, cardiomegaly, effusion, infiltration, mass, nodule, pneumonia, pneumothorax, consolidation, edema, emphysema, fibrosis, pleural thickening, and hernia. Previous works~\citep{docquier2012globalization} show that most pneumothorax patients have a spurious correlation with the chest drains. Chest drains are used to treat positive Pneumothorax cases. We adopt the strategy discussed in~\citet{murali2023beyond} to annotate chest drains for each sample. We use the official train/val/test split~\citep{wang2017chestx}.

\subsubsection*{RSNA breast mammograms}
The \textbf{RSNA-Mammo} dataset\footnote{\url{https://www.kaggle.com/competitions/rsna-breast-cancer-detection}} is a publicly available dataset containing 2D mammograms from 11,913 patients, with 486 diagnosed cancer cases. The task is to classify malignant cases from screening mammograms. We use a 70/20/10 train/validation/test split for evaluation as~\citet{ghosh2024mammo}.

\subsubsection*{VinDr breast mammograms}
The \textbf{VinDr-Mammo} dataset\footnote{\url{https://www.physionet.org/content/vindr-mammo/1.0.0/}}~\citep{nguyen2023vindr} is a publicly available 2D mammogram dataset of 5,000 exams (20,000 images) from Vietnam, each with four views. It includes breast-level BI-RADS assessment categories (1-5), breast density categories (A-D), and annotations for mammographic attributes (\eg mass, calcifications). Following~\citet{wen2024breast}, we classify patients with BI-RADS scores between 1 and 3 as normal and those with scores of 4 and 5 as abnormal. We adopt the train-test split from~\citet{nguyen2023vindr}.

\subsection{ Extended details on slice discovery algorithms}
\label{appendix_slice_discovery}

\textbf{Domino.}
Domino~\citep{eyuboglu2022domino} identifies systematic errors in machine learning models by leveraging cross-modal embeddings. It operates in three main steps: embedding, slicing, and describing.

\begin{enumerate}
    \item \textbf{Embedding}: Domino uses cross-modal models (\eg CLIP) to embed inputs and text in the same latent space. This enables the incorporation of semantic meaning from text into input embeddings, which is crucial for identifying coherent slices.
    
    \item \textbf{Slicing}: It employs an error-aware mixture model to detect underperforming regions within the embedding space. This model clusters the data based on embeddings, class labels, and model predictions to pinpoint areas where the model performance is subpar. The mixture model ensures that identified slices are coherent and relevant to model errors.
    
    \item \textbf{Describing}: Domino generates natural language descriptions for the discovered slices. It creates prototype embeddings for each slice and matches them with text embeddings to describe the common characteristics of the slice. This step provides interpretable insights into why the model fails on these slices.
\end{enumerate}

Domino's approach improves slice coherence and generates meaningful slice descriptions.

\textbf{Facts.} Facts~\citep{yenamandra2023facts} (First Amplify Correlations and Then Slice) aims to identify bias-conflicting slices in datasets through a two-stage process:

\begin{enumerate}
    \item \textbf{Amplify Correlations}: This stage involves training a model with a high regularization term to amplify its reliance on spurious correlations present in the dataset. This step helps segregate biased-aligned from bias-conflicting samples by making the model fit a simpler, biased-aligned hypothesis.
    
    \item \textbf{Correlation-aware Slicing}: In this stage, FACTS uses clustering techniques on the bias-amplified feature space to discover bias-conflicting slices. The method identifies subgroups where the spurious correlations do not hold, highlighting areas where the model underperforms due to these biases.
\end{enumerate}

Facts leverages a combination of bias amplification and clustering to reveal underperforming data slices, providing a foundation for understanding and mitigating systematic biases in machine learning models.

\subsection{ Extended details on error mitigation baselines}
\label{appendix_rr_mitigation}
We categorize the various bias mitigation algorithms and provide detailed descriptions for each category below.

\subsubsection*{Vanilla}
The empirical risk minimization (ERM) algorithm, introduced by Vapnik~\citep{Vapnik1999}, seeks to minimize the cumulative error across all samples.

\subsubsection*{Subgroup Robust Methods}
\textbf{GroupDRO:} GroupDRO~\citep{Sagawa2020} propose Group Distributionally Robust Optimization (GroupDRO), which enhances ERM by prioritizing groups with higher error rates.
 \textbf{CVaRDRO:} Duchi and Namkoong \citep{Duchi2018} introduce a variant of GroupDRO that dynamically assigns weights to data samples with the highest losses.
 \textbf{LfF:} LfF~\citep{Nam2020} concurrently trains two models: the first model is biased, and the second is de-biased by re-weighting the loss gradient.
 \textbf{Just Train Twice (JTT):} JTT~\citep{Liu2021} propose an approach that initially trains an ERM model to identify minority groups in the training set, followed by a second ERM model where the identified samples are re-weighted.
 \textbf{LISA:} LISA~\citep{Yao2022} utilizes invariant predictors through data interpolation within and across attributes.
 \textbf{Deep Feature Re-weighting (DFR):} DFR~\citep{kirichenko2022last} suggests first training an ERM model and then retraining the final layer using a balanced validation set with group annotations.

\subsubsection*{Data Augmentation}
\textbf{Mixup:} Mixup~\citep{Zhang2018} proposes an approach that performs ERM on linear interpolations of randomly sampled training examples and their corresponding labels.

\subsubsection*{Domain-Invariant Representation Learning}
\textbf{Invariant Risk Minimization (IRM):} IRM~\citep{Arjovsky2019} learns a feature representation such that the optimal linear classifier on this representation is consistent across different domains.
\textbf{Maximum Mean Discrepancy (MMD):} MMD~\citep{Li2018} aims to match feature distributions across domains.
\textbf{Note: All methods in this category necessitate group annotations during training}.

\subsubsection*{Imbalanced Learning}
 \textbf{Focal Loss (Focal):} Focal~\citep{Lin2017} introduces Focal Loss, which reduces the loss for well-classified samples and emphasizes difficult samples.
 \textbf{Class-Balanced Loss (CBLoss):} CBLoss~\citep{Cui2019} suggests re-weighting by the inverse effective number of samples.
 \textbf{LDAM Loss (LDAM):} LDAM~\citep{Cao2019} employs a modified margin loss that preferentially weights minority samples.
 \textbf{Classifier Re-training (CRT):} CRT~\citep{Kang2020} decomposes representation learning and classifier training into two distinct stages, re-weighting the classifier using class-balanced sampling during the second stage.
 \textbf{ReWeightCRT:} ReWeightCRT~\citep{Kang2020} proposes a re-weighted variant of CRT.

\subsection{ Learning Projection from classifier to VLR space}
\label{appendix: projection}
$\pi$ is a learnable projection function, $\pi: \Phi \rightarrow \Psi^I$, projecting the image representation of the classifier $\Phi(x)$ to the VLR space, $\Psi(x)$, where $x\in\mathcal{D}_{train}$. $\mathcal{D}_{train}$ denotes the training set. We follow~\citep{moayeri2023text} to learn $\pi$. Specifically, $\pi$ is an affine transformation, \ie $\pi_{W, b}(z) = W^Tz+b$, where $W$ and $b$ are the learnable weights and biases of the projector $\pi$. To retain the original semantics in the classifier representation space, we optimize the following objective:
{\small
\begin{align}
    W, b = \arg \min_{W, b} \frac{1}{|\mathcal{D}_{\text{train}}|} 
    \sum_{x \in \mathcal{D}_{\text{train}}}
    \big\| W^T\Phi(x) + b - \Psi(x) \big\|_2^2
\end{align}
}

\subsection{ Precision@k}
\label{appendix: precision@k}
\texttt{\textbf{Precision@k}}~\cite{eyuboglu2022domino, yenamandra2023facts} measures the degree to which the predicted slices overlap with the ground truth slices in a dataset.

Let $S = \{s_1, s_2, \dots, s_l\}$ represent the ground truth bias-conflicting slices in a dataset $\mathcal{D}$. A slice discovery algorithm $A$ predicts a set of slices $\hat{S} = \{\hat{s}_1, \hat{s}_2, \dots, \hat{s}_m\}$. For each predicted slice $\hat{s}_j$, let $O_j = \{o_{j1}, o_{j2}, \dots, o_{jn}\}$ denote the sequence of sample indices ordered by the decreasing likelihood that each sample belongs to the predicted slice $\hat{s}_j$.

Given a ground truth slice $s_i$ and a predicted slice $\hat{s}_j$, we compute their similarity as:
\[
P_k(s_i, \hat{s}_j) = \frac{1}{k} \sum_{i=1}^{k} \mathbb{I}[x_{o_{ji}} \in s_i],
\]
where $P_k(s_i, \hat{s}_j)$ is the proportion of the top $k$ samples in the predicted slice $\hat{s}_j$ that overlap with the samples in the ground truth slice $s_i$, and $\mathbb{I}$ is an indicator function that returns 1 if the sample belongs to $s_i$ and 0 otherwise.

For each ground truth slice $s_i$, we map it to the most similar predicted slice $\hat{s}_j$ by maximizing $P_k(s_i, \hat{s}_j)$. We then compute the average similarity score between the ground truth slices and their best-matching predicted slices. Specifically, the \texttt{Precision@k} for a slice discovery algorithm $A$ is given by:
\[
\texttt{Precision@k}(A) = \frac{1}{l} \sum_{i=1}^{l} \max_{j \in [m]} P_k(s_i, \hat{s}_j),
\]
where $l$ is the number of ground truth slices, $m$ is the number of predicted slices, and $P_k(s_i, \hat{s}_j)$ is the similarity score for the ground truth slice $s_i$ and predicted slice $\hat{s}_j$.

This metric evaluates how well the algorithm's predicted slices match the bias-conflicting slices in the dataset, with higher scores indicating better alignment between predicted and ground truth slices. By computing the \texttt{Precision@k}, we can assess the effectiveness of slice discovery algorithms in identifying and isolating the most significant bias-conflicting regions in the data.


\subsection{ Clip Score}
\label{appendix: clip_score_def}
\citet{kim2024discovering} introduces the CLIP score, a metric that leverages the similarity between language and vision embeddings to quantify the influence of specific attributes on misclassified samples. In their method, attributes frequently present in misclassified images receive a high CLIP score, while absent ones score lower. For instance, in the Waterbirds dataset, the CLIP score for "bamboo" is high, as many misclassified waterbirds appear with bamboo in the background.

We propose a modification to the CLIP score. As discussed in Sec.~\ref{sub:sent_retrieval}, our goal is to identify visual attributes that are prevalent in correctly classified samples but absent in misclassified ones. This approach provides deeper insights into the attributes contributing to correct classifications, which is particularly valuable for medical images. In scenarios such as pneumothorax detection in the NIH dataset, understanding biases incorrectly classified cases—such as the presence of chest tubes—can help isolate features that lead to reliable diagnoses while addressing spurious correlations. Formally we define the CLIP score corresponding to the attribute \texttt{attr} and a dataset $\mathcal{D}$ as,

{\scriptsize
\begin{align*}
    s_{CLIP}(\texttt{attr}, \mathcal{D}) = \texttt{sim}(\texttt{attr}, \mathcal{D}_{correct}) - \texttt{sim}(\texttt{attr}, \mathcal{D}_{wrong}) ,
\end{align*}
}
where \texttt{attr} is the attribute obtained from the specific hypothesis by LLM, described in Sec.~\ref{sec:hypthosis}, $\mathcal{D}_{correct}$ and $\mathcal{D}_{wrong}$ are the correctly classified and misclassified samples. Also, $\texttt{sim}(\texttt{attr}, \mathcal{D})$ is the similarity between the attribute \texttt{attr} and the dataset $\mathcal{D}$, estimated as the average cosine similarity between normalized embedding of a word $\Psi^T(\texttt{attr})$ and images $\Psi^I(\texttt{x})$ for $x\in \mathcal{D}$, where
\begin{align*}
    \texttt{sim}(\texttt{attr}, \mathcal{D})=\frac{1}{\mathcal{D}}\sum_{x\in\mathcal{D}}\Psi^I(x)\Psi^T(\texttt{attr})
\end{align*}
Refer to Appendix~\ref{clip_score} for the results.

\subsection{ Prompts used by LLM for hypotheses generation}
\label{appendix: prompts}
The following is a general template of the prompt utilized to generate the hypotheses from LLM, discussed in Sec.~\ref{sec:hypthosis}. 
In this template, we substitute the <task> placeholders with bird species, hair color, animal species, pneumothorax, cancer, and abnormality based on the corresponding dataset -- Waterbirds, CelebA, MetaShift, NIH, RSNA-Mammo, and VinDr-Mammo. The modalities are natural images, chest-x-rays, and 2D mammograms. \textbf{Crucially, we only replace these two placeholders. We never include the actual dataset names or words like ``water'', ``land'', ``gender'', ``tube'', ``background'' or any other attributes leading to model's mistakes in the prompt, as these may bias the LLM's output}. For medical images, we also add: \texttt{Ignore `\_\_\_' as they are due to anonymization. We focus only on positive <disease> patients}, as many reports consist of `\_\_\_' for clarity. top <K> depends on the dataset discussed in the experiment section (Sec.~\ref{sub:experimental_details}).

\begin{table*}[h]
\centering
\begin{tcolorbox}[colback=white, colframe=black, title=Prompt for Hypothesis Generation, fonttitle=\bfseries, width=\textwidth]
\sloppy
\textbf{Context:} \texttt{<task>} classification from \texttt{<modality>} using a deep neural network. \\[0.5em]
\textbf{Analysis Post-Training:} On a validation set:
\begin{enumerate}[label=\alph*.]
    \item Get the difference between the image embeddings of correct and incorrectly classified samples to estimate the features present in the correctly classified samples but missing in the misclassified samples.
    \item Retrieve the top \texttt{<K>} sentences from the <captions/radiology report> that match closely to the embedding difference in step a.
    \item The sentence list is given below:
\end{enumerate}

\begin{tcolorbox}[colback=gray!10!white, colframe=gray!75!black, title=\texttt{TopK} Sentence List, width=\textwidth]
\textbf{\color{black} Retrieved using Sec.~\ref{sub:sent_retrieval}}
\end{tcolorbox}

These sentences represent the features present in the correctly classified samples but missing in the misclassified samples.

\textbf{Task:} Consider the consistent attributes present in the descriptions of correctly classified and misclassified samples regarding \texttt{<task>}. Formulate hypotheses based on these attributes. Attributes include all concepts (e.g., explicit or implicit anatomies, observations, symptoms of change related to the disease, concepts leading to potential bias in medical images, or visual cues in natural images) in the sentences. Assess how these characteristics might influence the classifier's performance.

Your response should only contain the list of top hypotheses, formatted as follows:
\noindent
\begin{lstlisting}[style=mypython]
hypothesis_dict = {
    'H1': 'The classifier is making mistake as it is biased toward <attribute>',
    'H2': 'The classifier is making mistake as it is biased toward <attribute>',
    'H3': 'The classifier is making mistake as it is biased toward <attribute>',
    ...
}
\end{lstlisting}


To effectively test Hypothesis 1 (H1) using the CLIP language encoder, create prompts explicitly validating H1. These prompts will help generate text embeddings that capture the essence of the hypothesis, which can be compared with the image embeddings from the dataset. The goal is to verify alignment with or violation of H1. Prompts must focus only on the \texttt{<task>}. Each hypothesis must have five prompts, formatted as:

\noindent
\begin{lstlisting}[style=mypython]
prompt_dict = {
    'H1_<attribute>': [List of prompts],
    'H2_<attribute>': [List of prompts],
    ...
}
\end{lstlisting}


Final response format strictly:
\noindent
\begin{lstlisting}[style=mypython]
hypothesis_dict 
prompt_dict 
\end{lstlisting}

\end{tcolorbox}
\caption{Detailed description of the prompt for hypothesis generation and analysis for the \texttt{<task>} classification problem.}
\label{table:prompt_hypothesis}
\end{table*}

\clearpage

\subsection{ Prompts and details on the experiments in RQ5 with instruction-tuned models (\eg LLaVA)}
\label{appendix: llava_med_prompt}
In this setup, we don't use CLIP as VLR for the retrieval step discussed in Sec.~\ref{sub:sent_retrieval}. Instead, using the instruction-tuned vision language models (LLaVA-1.5 7B for natural images; cheXagent and RaDialog for CXRs), we first select the correctly classified images by the classifier $f$. Next, for each of the images, we pass them through the vision encoder in LLaVA and use the prompt for the natural images: ``\texttt{Describe the image}'' for the language model in LLaVA. For NIH, we use the prompt: 
\lstset{
    basicstyle=\ttfamily\small,
    breaklines=true,
    breakatwhitespace=true,
    columns=fullflexible,
    numbers=none 
}
\begin{lstlisting}[basicstyle=\ttfamily\footnotesize, frame=single, xleftmargin=0pt, framexleftmargin=0pt]
You are a radiologist. Based on the provided Chest X-Ray image and generate a structured report. The report should include sections for `Findings,' `Impression,' and `Recommendations,' emphasizing relevant findings like consolidation, effusion, cardiomegaly, pneumonia, or pneumothorax. Use a formal radiology reporting style.
\end{lstlisting}

We select the texts for all the correctly classified images and follow\ladder's pipeline discussed in Sec~\ref{sec:hypthosis} to generate the hypothesis (results shown in Fig.~\ref{fig:llava-heatmap}). Finally, we utilize\ladder's mitigation strategy, discussed in Sec.~\ref{sub: bias_mitigation} to mitigate the biases (results shown in Tab.~\ref{fig:llava-precision-mit}). \textbf{Note: in this experiment, we did not use any language explicitly. However, there is always a trade-off between getting language or using an instruction-tuning model like LLaVA.}

\subsection{Prompts and examples of metadata for detecting biases beyond radiology reports in RQ6}
\label{appendix: metadatabias}
Refer to Tab.~\ref{table:metadat_bias} for the prompt and the example of Python dictionary of metadata details of the correctly classified cancer patients to detect biases using~\ladder.

\begin{table*}[h]
\centering
\begin{tcolorbox}[colback=white, colframe=black, fonttitle=\bfseries, width=\textwidth]
\sloppy
\textbf{Context:} 
Breast cancer classification from mammograms using a deep neural network

\textbf{Analysis post-training:} On a validation set, you are provided with the metadata details for the correctly classified positive cancer patients in a Python dictionary, as follows

\begin{tcolorbox}[colback=gray!10!white, colframe=gray!75!black, title=Metadata Dictionary (Sample Entries), fonttitle=\bfseries, width=\textwidth]
\texttt{
\begin{itemize}
    \item \textbf{Patient 1:}  
        \{site\_id: 1, laterality: L, view: MLO, age: 71, biopsy: 1,  
        invasive: 1, BIRADS: 0, implant: 0, density: B, machine\_id: 49,  
        photometric\_interpretations: Monochrome 1,  
        voi\_lut\_function: SIGMOID, pixel\_intensity\_relationship: LOG\}
    \item \textbf{Patient 2:}  
        \{site\_id: 2, laterality: L, view: CC, age: 83, biopsy: 0,  
        invasive: 0, BIRADS: 0, implant: 1, density: D, machine\_id: 49,  
        photometric\_interpretations: Monochrome 1,  
        voi\_lut\_function: SIGMOID, pixel\_intensity\_relationship: LOG\}
    \item $\cdots$ (Additional metadata entries omitted for brevity)
\end{itemize}
}
\end{tcolorbox}

\textbf{Task:}
    Consider the consistent attributes present in the dictionary regarding the positive cancer patients. Formulate hypotheses based on these attributes. Assess how these characteristics might be influencing the classifier's performance. Your response should contain only the list of top hypothesis, nothing else. For the response, you should be the following python dictionary template, no extra sentence:
    
    \noindent

\begin{lstlisting}[style=mypython]
hypothesis_dict = {
    'H1': 'The classifier is making mistake as it is biased toward <attribute>',
    'H2': 'The classifier is making mistake as it is biased toward <attribute>',
    'H3': 'The classifier is making mistake as it is biased toward <attribute>',
    ...
}
\end{lstlisting}

    
\end{tcolorbox}
\caption{Prompts and examples of metadata for detecting biases beyond radiology reports in the experiment RQ6.}
\label{table:metadat_bias}
\end{table*}

\subsection{ Extended details on general experiments}
\label{appendix_experiments}

\subsubsection{Implementation details of the source model $f$ using ERM}
\label{app:implement}

For natural images and chest X-rays (CXRs), we resize the images to 224$\times$224 and train ResNet-50 (RN)\citep{he2016deep} and Vision Transformer (ViT)\citep{dosovitskiy2020image} models as $f$ to predict labels. We explore various pretraining methods for initializing model weights, including supervised learning (Sup), SIMCLR\citep{chen2020simple}, Barlow Twins~\citep{zbontar2021barlow}, DINO~\citep{caron2021emerging}, and CLIP-based pretraining~\citep{radford2021learning}. The pretraining datasets utilized include ImageNet-1K (IN1)\citep{deng2009imagenet}, ImageNet-21K (IN-21K)\citep{ridnik2021imagenet}, SWAG~\citep{singh2022revisiting}, LAION-2B~\citep{schuhmann2022laion}, and OpenAI-CLIP (OAI)~\citep{radford2021learning}. For instance,``RN Sup IN1k'' refers to a ResNet model pretrained using supervised learning and ImageNet-1K.

We train both ResNet and ViT models as $f$ for natural images and NIH-CXR following the setup in~\citet{yang2023change}\footnote{\url{https://github.com/YyzHarry/SubpopBench}}. Preprocessing steps include resizing the images to 224$\times$224, applying center-cropping, and normalizing the images using ImageNet channel statistics. Consistent with prior work~\citep{Idrissi2022, Imazilov2022}, we apply stochastic gradient descent (SGD) with momentum for optimization across all image datasets.
Each model is trained for a total of 30,000 steps across all datasets, with specific training on Waterbirds and MetaShift for 5,000 steps each. For NIH, we utilize the Adam optimizer with a learning rate of 0.0001 and train for 60 epochs to achieve optimal convergence.

For RSNA-Mammo, we leverage the setting from one of the leading Kaggle competition solutions\footnote{\url{https://github.com/Masaaaato/RSNABreast7thPlace}}. In this setup, the images are resized to 1520$\times$912, and we train an EfficientNet-B5 model~\citep{tan2019efficientnet} for 9 epochs using the SGD optimizer, with a learning rate of 5e-5 and a weight decay of 1e-4.

Additionally, for CXR-CLIP, we use their pretrained models\footnote{\url{https://github.com/kakaobrain/cxr-clip}}, which were trained on MIMIC-CXR and CheXpert (MC) datasets. For Mammo-CLIP, we utilize their EN-B5 variant\footnote{\url{https://huggingface.co/shawn24/Mammo-CLIP/blob/main/Pre-trained-checkpoints/b5-model-best-epoch-7.tar}}.

\subsubsection{Ablations}
For the captioning ablations, we compare the performance of\ladder using BLIP~\citep{li2022blip}, BLIP-2~\citep{li2023blip}, ClipCap~\citep{mokady2021clipcap}, and GPT-4o~\citep{wu2024gpt}. Additionally, for LLMs, we compare the performance of\ladder with GPT-4o~\citep{wu2024gpt}, Claude 3.5 Sonnet, Llama 3.1 70B~\citep{dubey2024llama}, and Gemini 1.5 Pro~\citep{team2024gemini}.

\subsubsection{Radiology text synthesis for 2D Mammograms}
\label{app: rad_report_mammo}
In~\citet{ghosh2024mammo}, the authors generate mammography reports using labeled mammographic attributes from the VinDr dataset in collaboration with a board-certified radiologist. This approach leverages the templated nature of breast mammogram reports, which are more standardized than those for other medical imaging modalities. This standardized structure follows protocols like BI-RADS (Breast Imaging-Reporting and Data System), which promotes uniformity in reporting~\citep{palanisamy2023reporting}. Specifically, they focus on the following attributes: \texttt{mass}, \texttt{architectural distortion}, \texttt{calcification}, \texttt{asymmetry (focal, global)}, \texttt{density}, \texttt{suspicious lymph nodes}, \texttt{nipple retraction}, \texttt{skin retraction}, and \texttt{skin thickening}. Then they follow the report templates with radiologist-defined prompts in~\citet{ghosh2024mammo}, describing key parameters such as:

\textbf{Attribute Value}: Positive, negative, \etc

\textbf{Subtype}: Suspicious, obscured, spiculated, \etc

\textbf{Laterality}: Left or right breast.

\textbf{Position}: Upper, lower, inner, outer quadrant.

\textbf{Depth}: Anterior, mid, or posterior.

Finally, they generate concise report-like sentences by substituting these values into the templates. The authors leverage these sentences in Mammo-FActOR to perform weakly supervised localization of mammographic findings. In our work, we collect all these sentences to probe the EN-B5 classifier $f$, analyzing its errors during the retrieval step (Sec.~\ref{sub:sent_retrieval}) for the RSNA-Mammo and VinDr-Mammo datasets.

Below are some examples of mammography report sentences corresponding to the specific mammographic attributes.

\textbf{Mass:}
\lstset{
    basicstyle=\ttfamily\small,
    breaklines=true,
    breakatwhitespace=true,
    columns=fullflexible,
    numbers=none 
}
\begin{lstlisting}
    1. there is a mass in the right breast
    2. there is a mass in the right breast at anterior depth
    3. there is a mass in the upper right breast at mid-depth
                       ...
\end{lstlisting}

\textbf{Architectural distortion:}
\begin{lstlisting}
    1. there is architectural distortion in the right breast
    2. there is architectural distortion in the right breast at anterior depth
    3. there is architectural distortion in the right breast at mid-depth
                       ...
\end{lstlisting}

\textbf{Calcification:}
\begin{lstlisting}
    1. there is calcification in the right breast
    2. there is calcification in the right breast at anterior depth
    3. there is calcification in the right breast at mid depth
                       ...
\end{lstlisting}

\textbf{Asymmetry:}
\begin{lstlisting}
    1. there is a developing asymmetry in the outer right breast
    2. there is an asymmetry in the inner right breast at anterior depth
    3 .there is an asymmetry in the right breast at mid-depth
                       ...
\end{lstlisting}

\textbf{Global Asymmetry:}
\begin{lstlisting}
    1. there is a global asymmetry in the right breast
    2. there is a new global asymmetry in the right breast
    3. there is a global asymmetry in the inner right breast
                       ...
\end{lstlisting}

\textbf{Focal Asymmetry:}
\begin{lstlisting}
    1. mthere is a focal asymmetry in the right breast
    2. there is a focal asymmetry in the right breast at anterior depth
    3. there is a focal asymmetry in the right breast at mid depth
                       ...
\end{lstlisting}

\textbf{Density:}
\begin{lstlisting}
    1. the breasts being almost entirely fatty
    2. scattered areas of fibroglandular density
    3. the breast tissue is heterogeneously dense
    4. the breasts are extremely dense
                       ...
\end{lstlisting}

\textbf{Suspicious lymph node:}
\begin{lstlisting}
    1. there is a suspicious lymph node in the  right axilla
    2. there is a hyperdense lymph node in the  right axillary tail
    3. there is an increased lymph node in the  right axillary tail
                       ...
\end{lstlisting}

\textbf{Suspicious lymph node:}
\begin{lstlisting}
    1. there is a suspicious lymph node in the  right axilla
    2. there is a hyperdense lymph node in the  right axillary tail
    3. there is an increased lymph node in the  right axillary tail
                       ...
\end{lstlisting}

\textbf{Nipple retraction:}
\begin{lstlisting}
    1. there is a new nipple retraction in the right breast
    2. there is an increased nipple retraction in the right breast
    3. there is a possible nipple retraction in the right breast
                       ...
\end{lstlisting}

\textbf{Skin retraction:}
\begin{lstlisting}
    1. there is  skin retraction in the right breast
    2. there is  skin retraction in the inner right breast
    3. there is  skin retraction in the lower right breast
                       ...
\end{lstlisting}

\textbf{Skin thickening:}
\begin{lstlisting}
    1. there is increasing skin thickening of the periareolar right breast
    2. there is asymmetric skin thickening of the lower right breast
    3. there is asymmetric skin thickening of the inner right breast
                       ...
\end{lstlisting}

\subsection{Toy dataset construction}
\label{appendix_dataset_toy}
We construct a synthetic dataset based on the \textbf{CUB-200-2011}~\citep{Wah2011} dataset, classifying bird species into two categories: \textbf{Class 0} ($y = 0$) and \textbf{Class 1} ($y = 1$). Class 1 consists of the following bird species: \textit{Albatross}, \textit{Auklet}, \textit{Cormorant}, \textit{Frigatebird}, \textit{Fulmar}, \textit{Gull}, \textit{Jaeger}, \textit{Kittiwake}, \textit{Pelican}, \textit{Puffin}, \textit{Tern}, \textit{Gadwall}, \textit{Grebe}, \textit{Mallard}, \textit{Merganser}, \textit{Guillemot}, and \textit{Pacific Loon}. All remaining bird species are assigned to Class 0.
To introduce spurious correlations, we overlay two 3D boxes on each image. In the training set for Class 0, the majority of samples (95\%) were biased, with the yellow box consistently placed to the left of the red box. For Class 1, the boxes were randomly placed, introducing variability in their positioning.  In the validation and test sets, we split the positioning evenly, with 50\%  biased and 50\% random samples across both classes, ensuring a balanced evaluation of the model's reliance on spurious cues.

The primary goal of this dataset is to introduce a form of \textit{reasoning} beyond the mere presence or absence of spurious correlations. Unlike prior datasets that rely on background cues (\eg Waterbirds or Metashift) or attributes like gender (\eg CelebA), our dataset integrates positional reasoning. Specifically, for Class 0, the yellow box is consistently placed to the left of the red box, creating a spurious correlation. For Class 1, the boxes are randomly positioned, removing this shortcut. The relative positioning of the boxes allows the captions to encode spatial relationships, which can be consumed by large language models (LLMs) to reason about these spatial cues. We train an ImageNet pretrained-ResNet model (RN Sup IN1k) on this dataset. Predictably, the classifier latches onto the spurious correlation of rectangle position, leading to underperformance on subsets where the shortcut is absent. The model achieves a mean accuracy of 85.6\% and a worst-group accuracy (WGA) of 65.2\%. 

\begin{figure*}[h]
\centering
\includegraphics[width=0.7\textwidth]{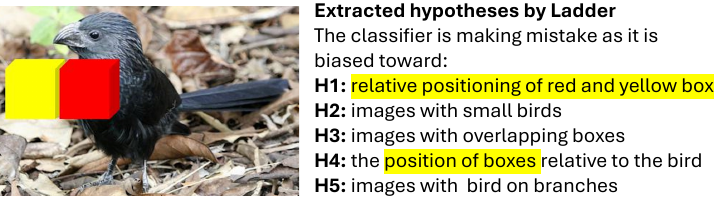}
\caption{Sample images of our toy dataset to validate the reasoning of LLM utilized by\ladder. 
The dataset has two classes. Images with class 0 are biased, with the yellow box always placed to the left of the red box. For images with class 1, the boxes are randomly placed.}
\label{fig:toy}
\end{figure*}

To analyze the model’s errors, we generate a corpus of rich captions for the validation set using a GPT-4o-based captioner. These captions describe both the presence of the rectangle and its position relative to the bird. Using\ladder, we aim to detect the reason for the classifier's mistakes and mitigate it.\ladder leverages the reasoning capabilities of LLMs to capture both the presence of the rectangles and their relative spatial position. In contrast, methods \eg PRIME, rely on external tagging models, which only detect the presence or absence of shortcuts. Furthermore, since \ladder discovers biased attributes via LLM-generated reasoning, it can effectively mitigate these biases without requiring ground truth annotations or prior knowledge of the attributes.

The data is split into training, validation, and test sets, with all metadata (including labels, rectangle positions) saved for future analysis.

\subsection{ Extended main results}
\label{appendix_results}





\subsubsection {Results on WGA for using all slice discovery methods:}
\label{appendix: domino-facts-ladder}
\begin{figure*}[h]
\begin{center}
\includegraphics[scale=0.6]{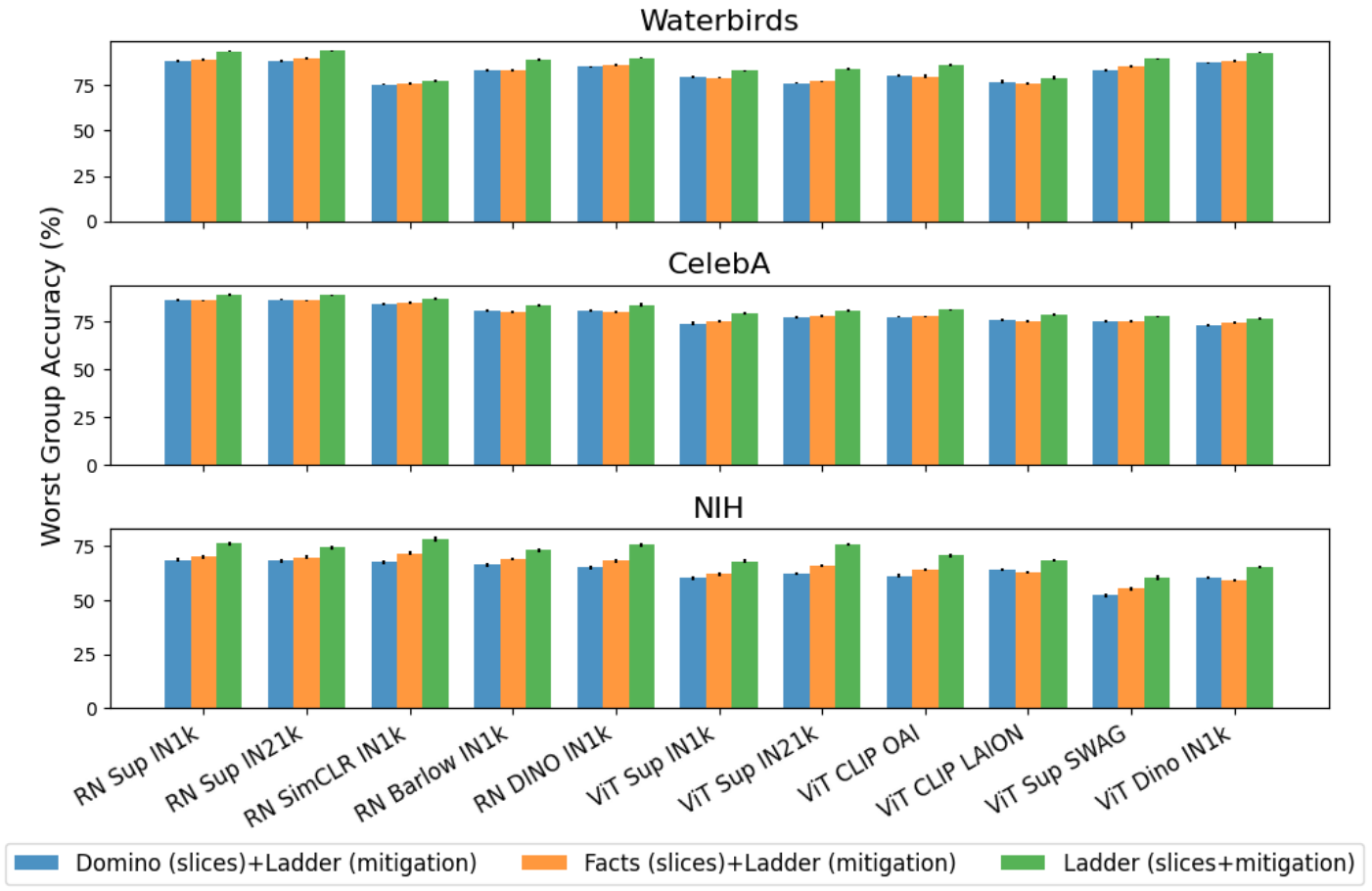}
\caption{\ladder slices consistently outperform those from Domino and Facts when combined with \ladder's bias mitigation strategy across various settings.} 
\label{fig:domino_compare_ladder_wga}
\end{center}
\end{figure*}

\begin{figure}[h]
\begin{center}
\includegraphics[width=\linewidth]{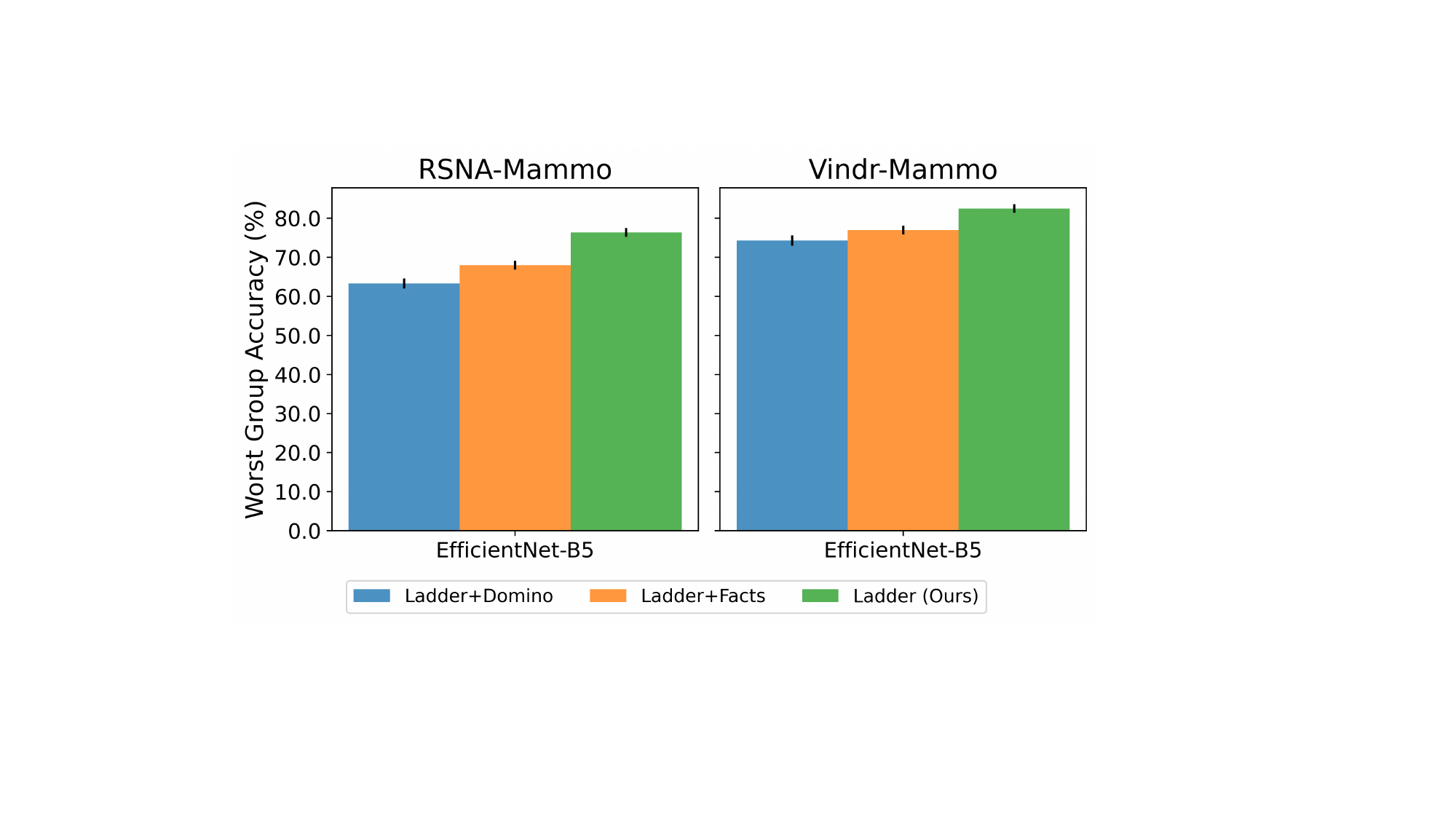}
\caption{\ladder improves WGA compared to other bias mitigation methods for RSNA-Mammo and VinDr-Mammo datasets.} 
\label{fig:domino-compare-breast}
\end{center}
\end{figure}

Fig.\ref{fig:domino_compare_ladder_wga} shows that\ladder improves WGA compared to other slice discovery methods for natural images and CXRs. In this experimental setup, we first discover the slices with Domino~\cite{eyuboglu2022domino}, Facts~\cite{yenamandra2023facts} and\ladder's hypothesis-driven approaches. Next, we apply\ladder's mitigation approach for each discovered slice to mitigate the biases and compute the WGA for each slice discovery method. As~\ladder detects the slices precisely, it achieves better WGA compared to Domino and Facts. Fig.~\ref{fig:domino-compare-breast} shows\ladder improves WGA compared to other slice discovery methods for RSNA-Mammo and VinDr-Mammo datasets.

\begin{figure}[h]
\begin{center}
\includegraphics[width=0.8\linewidth]{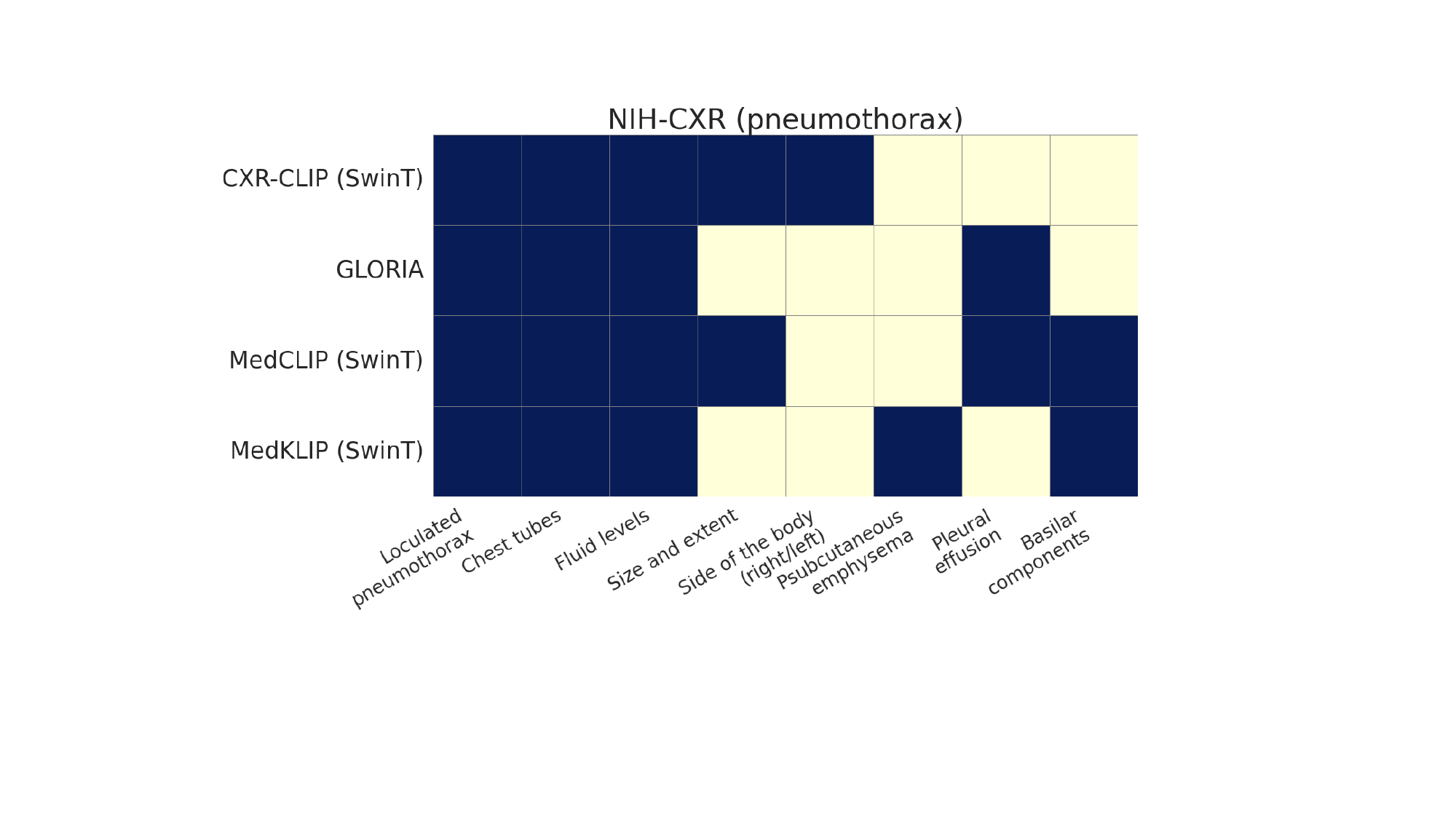}
\caption{Effect of different VLRs for CXRs on biased attribute discovery by\ladder. Bright/light colors denote presence/absence of the attributes.}
\label{fig:cxr_ablations}
\end{center}
\vspace{-1.1em}
\end{figure}

\subsubsection{Closest hypothesis to the ground truth attribute}
Tab.~\ref{tab:cnn_attrs} and Tab.~\ref{tab:vit_attrs} show the \texttt{top3} hypotheses for RN Sup IN1K (convolution-based) and ViT Sup IN1K (transformer-based) architectures, respectively. These hypotheses are the most similar to the ground truth attribute on which the source model $f$ is biased.
\label{appendix: closest_attr}
\begin{table}[h]
\centering
\caption{Top 3 associated hypotheses for the ground truth biased attribute for ViT Sup IN1K model on various datasets}
\label{tab:vit_attrs}
\setlength{\tabcolsep}{5pt}
\begin{center}
\resizebox{\linewidth}{!}{
\begin{tabular}{lcc}
\toprule
\small \textbf{Dataset (Label)} & 
\small \textbf{Attribute} &  
\small \textbf{\makecell{Top 3 hypotheses}}  \\
\midrule
\small Waterbirds (waterbird) & \small Water  & 
\small \begin{minipage}{0.4\textwidth}
1. activities like swimming or flying \\
2. conditions like cloudy or sunny \\
3. presence of objects like boats or rocks
\end{minipage} \\

\midrule
\small Waterbirds (landbird) & \small Land & 
\small \begin{minipage}{0.4\textwidth}
1. bird in the middle of a forest \\
2. yellow bird \\ 
3. bird sitting on top of a tree
\end{minipage} \\

\midrule
\small CelebA (Blonde) & \small Women & 
\small \begin{minipage}{0.4\textwidth}
1. woman wearing red dress \\ 
2. woman with red top \\
3. black jacket
\end{minipage} \\

\midrule
\small MetaShift (Dog) & \small Outdoor & 
\small \begin{minipage}{0.4\textwidth}
1. presence of a leash \\
2. presence of a ball \\ 
3. presence of a car
\end{minipage} \\

\midrule
\small MetaShift (Cat) & \small Indoor & 
\small \begin{minipage}{0.4\textwidth}
1. beds\\
2. windows \\
3. televisions 
\end{minipage} \\

\bottomrule
\end{tabular}
}
\end{center}
\end{table}

\begin{table}[h]
\caption{Top 3 associated hypotheses for the ground truth biased attribute for RN Sup IN1K model on various datasets}
\label{tab:cnn_attrs}
\setlength{\tabcolsep}{5pt}
\begin{center}
\resizebox{\linewidth}{!}{
\begin{tabular}{lcc}
\toprule
\small \textbf{Dataset (Label)} & 
\small \textbf{Attribute} &  
\small \textbf{\makecell{Top 3 hypotheses}}  \\
\midrule
\small Waterbirds (waterbird) & \small Water  & 
\small \begin{minipage}{0.4\textwidth}
1. water bodies like oceans and lakes \\
2. actions like flying or sitting \\
3. conditions, e.g., cloudy skies
\end{minipage} \\

\midrule
\small Waterbirds (landbird) & \small Land & 
\small \begin{minipage}{0.4\textwidth}
1. bird being in flight \\
2. bird perching on top of a tree \\ 
3. bird perching on a tree branch 
\end{minipage} \\

\midrule
\small CelebA (Blonde) & \small Women & 
\small \begin{minipage}{0.4\textwidth}
1. woman with long hair \\ 
2. woman wearing red dress \\
3. a black jacket
\end{minipage} \\

\midrule
\small MetaShift (Dog) & \small Outdoor & 
\small \begin{minipage}{0.4\textwidth}
1. dogs in motion \\
2. dogs on leashes \\ 
3. beach environments 
\end{minipage} \\

\midrule
\small MetaShift (Cat) & \small Indoor & 
\small \begin{minipage}{0.4\textwidth}
1. televisions \\
2. windows \\
3. beds
\end{minipage} \\

\midrule
\small NIH (pneumothorax) & \small Chest tube & 
\small \begin{minipage}{0.4\textwidth}
1. the presence of chest tubes \\
2. loculated pneumothorax \\
3. size and extent of pneumothorax
\end{minipage} \\

\midrule
\small RSNA-Mammo (cancer) & \small Calcification & 
\small \begin{minipage}{0.4\textwidth}
1. scattered calcifications \\
2. vascular calcifications \\ 
3. bilateral occurrences
\end{minipage} \\

\bottomrule
\end{tabular}
}
\end{center}
\end{table}

\begin{table}[h]
\centering
\scriptsize 
\setlength{\tabcolsep}{4pt} 
\renewcommand{\arraystretch}{1.1} 
\caption{\textbf{Token Usage and Cost for Each LLM.} Each row shows the breakdown for an LLM extracting hypotheses across all 6 datasets, using RN Sup IN1k (natural images / CXRs) and EN-B5 (mammograms).}
\label{tab:llm_token_breakdown}
\begin{tabular}{lrrr}
\toprule
\textbf{Model Name} & \textbf{Input Tokens} & \textbf{Output Tokens} & \textbf{Total Cost} \\
\midrule
GPT-4o & 33,217 & 4,284 & \$2.51 \\
Claude 3.5 Sonnet & 34,888 & 4,473 & \$0.17 \\
Gemini 1.5 Pro & 33,872 & 4,378 & \$0.32 \\
Llama 3.1 70B & 32,688 & 4,176 & \$0.05 \\
\midrule
\textbf{Total} & 134,665 & 17,311 & \textbf{\$3.05} \\
\bottomrule
\end{tabular}
\end{table}


\begin{table*}[h!]
\footnotesize
\centering
\caption{Benchmarking error mitigation methods over 3 seeds for CNN models (EN-B5 for mammograms and RN Sup IN1k for the rest). For natural images (Waterbirds and CelebA), we report mean accuracy. For medical images (NIH, RSNA and VinDr), we report mean AUROC. We bold-face and underline the best and second-best results, respectively.}
\label{tab:app_mitigation_RN50}
\setlength{\tabcolsep}{3pt}
\begin{adjustbox}{max width=\textwidth}
\begin{tabular}{lcccccccccc}
\toprule
\multirow{2.5}{*}{\textbf{Method}} & 
\multicolumn{2}{c}{\textbf{Waterbirds}} 
& \multicolumn{2}{c}{\textbf{CelebA}} 
& \multicolumn{2}{c}{\textbf{NIH}} 
& \multicolumn{2}{c}{\textbf{RSNA}} 
& \multicolumn{2}{c}{\textbf{VinDr}}\\
\cmidrule(lr){2-3} \cmidrule(lr){4-5}
\cmidrule(lr){6-7} \cmidrule(lr){8-9}
\cmidrule(lr){9-10} \cmidrule(lr){10-11}
& \textbf{Mean(\%)} & \textbf{WGA(\%)} 
& \textbf{Mean(\%)} & \textbf{WGA(\%)}  
& \textbf{Mean(\%)} & \textbf{WGA(\%)}  
& \textbf{Mean(\%)} & \textbf{WGA(\%)}
& \textbf{Mean(\%)} & \textbf{WGA(\%)} \\

\midrule
Vanilla (ERM) 
& 88.2$_{\pm0.7}$ & 69.1$_{\pm1.2}$ 
& 94.1$_{\pm0.2}$ & 62.2$_{\pm1.5}$ 
& \textbf{87.4}$_{\pm0.0}$ & 60.3$_{\pm0.0}$ 
& \textbf{86.5}$_{\pm0.0}$ & 69.8$_{\pm0.0}$
& \textbf{86.9}$_{\pm0.0}$ & 45.6$_{\pm0.0}$ \\
\midrule

Mixup 
& 88.5$_{\pm0.5}$ & 77.3$_{\pm0.5}$
& \underline{94.5}$_{\pm0.1}$ & 57.8$_{\pm0.8}$  
& 85.1$_{\pm0.0}$ & 67.6$_{\pm0.8}$ 
& 84.5$_{\pm0.0}$ & 64.8$_{\pm0.0}$ 
& 83.2$_{\pm0.0}$ & {65.3$_{\pm0.0}$}\\

IRM 
& 88.1$_{\pm0.2}$ & 74.3$_{\pm0.1}$
& \underline{94.5}$_{\pm0.5}$ & 63.3$_{\pm2.5}$ 
& 83.2$_{\pm0.0}$ & 63.4$_{\pm0.0}$   
& 83.3$_{\pm0.0}$ & 68.4$_{\pm0.0}$ 
& 83.5$_{\pm0.0}$ & {65.2$_{\pm0.0}$}\\

MMD 
& 92.5$_{\pm0.1}$ & 83.5$_{\pm1.1}$
& 92.5$_{\pm0.6}$ & 22.7$_{\pm2.5}$
& 84.6$_{\pm0.0}$ & 65.4$_{\pm0.0}$   
& 84.2$_{\pm0.0}$ & 69.1$_{\pm0.0}$ 
& 81.2$_{\pm0.0}$ & {64.8$_{\pm0.0}$}\\
\midrule

Focal 
& 89.3$_{\pm0.2}$ & 71.6$_{\pm0.8}$
& \textbf{94.9}$_{\pm0.3}$ & 59.3$_{\pm2.0}$  
& 85.5$_{\pm0.0}$ & 68.9$_{\pm0.7}$  
& 83.6$_{\pm0.0}$ & 65.5$_{\pm0.0}$ 
& 82.6$_{\pm0.0}$ & {63.7$_{\pm0.0}$}\\

CBLoss 
& 91.3$_{\pm0.7}$ & 86.1$_{\pm0.3}$
& 91.2$_{\pm0.7}$ & 87.3$_{\pm0.5}$  
& 85.5$_{\pm0.0}$ & 63.4$_{\pm0.0}$   
& 83.2$_{\pm0.0}$ & 65.1$_{\pm0.0}$ 
& 81.7$_{\pm0.0}$ & {62.5$_{\pm0.0}$}\\

LDAM 
& 91.3$_{\pm0.7}$ & 86.1$_{\pm0.3}$
& \underline{94.5}$_{\pm0.2}$ & 58.3$_{\pm2.5}$ 
& 84.3$_{\pm0.0}$ & 69.4$_{\pm0.2}$   
& 81.6$_{\pm0.0}$ & 63.5$_{\pm0.0}$ 
& 81.2$_{\pm0.0}$ & {62.2$_{\pm0.0}$}\\

CRT 
& 90.5$_{\pm0.0}$ & 79.7$_{\pm0.3}$ 
& 92.5$_{\pm0.1}$ & 87.3$_{\pm0.3}$  
& 82.7$_{\pm0.0}$ & 68.5$_{\pm0.0}$   
& 82.7$_{\pm0.0}$ & 68.8$_{\pm0.0}$ 
& 82.9$_{\pm0.0}$ & {63.3$_{\pm0.0}$} \\

ReWeightCRT 
& 91.3$_{\pm0.1}$ & 78.4$_{\pm0.1}$
& 92.5$_{\pm0.2}$ & 87.2$_{\pm0.5}$ 
& 83.0$_{\pm0.0}$ & 69.5$_{\pm0.0}$   
& 82.4$_{\pm0.0}$ & 68.3$_{\pm0.0}$ 
& 82.9$_{\pm0.0}$ & {63.3$_{\pm0.0}$} \\

\midrule
JTT 
& 88.8$_{\pm0.7}$ & 84.5$_{\pm0.3}$ 
& 90.6$_{\pm2.2}$ & 87.2$_{\pm7.5}$  
& 85.1$_{\pm0.0}$ & 70.4$_{\pm0.0}$  
& 84.6$_{\pm0.0}$ & 68.5$_{\pm0.0}$ 
& 83.7$_{\pm0.0}$ & 66.1$_{\pm0.0}$ \\

GroupDRO 
& 88.8$_{\pm1.7}$ & 87.1$_{\pm1.3}$ 
& 91.4$_{\pm0.6}$ & \underline{88.1}$_{\pm0.7}$    
& 85.2$_{\pm0.0}$ & 71.1$_{\pm0.0}$
& 85.1$_{\pm0.0}$ & 72.3$_{\pm0.0}$ 
& 82.7$_{\pm0.0}$ & 67.1$_{\pm0.0}$ \\

CVaRDRO 
& 89.8$_{\pm0.4}$ & 85.4$_{\pm2.3}$ 
& \underline{94.5}$_{\pm0.1}$ & 83.1$_{\pm1.5}$  
& 85.7$_{\pm0.1}$ & 71.3$_{\pm0.0}$ 
& 85.4$_{\pm0.0}$ & 71.7$_{\pm0.0}$ 
& 82.7$_{\pm0.0}$ & 67.1$_{\pm0.0}$ \\

LfF 
& 87.0$_{\pm0.3}$ & 75.2$_{\pm0.7}$ 
& 81.1$_{\pm5.6}$ & 63.0$_{\pm4.4}$  
& 75.9$_{\pm0.0}$ & 61.6$_{\pm0.0}$ 
& 79.8$_{\pm0.0}$ & 66.4$_{\pm0.0}$ 
& 82.4$_{\pm0.0}$ & 64.5$_{\pm0.0}$ \\

LISA 
& 92.8$_{\pm0.3}$ & 88.7$_{\pm0.6}$  
& 92.6$_{\pm0.1}$ & 86.2$_{\pm1.1}$    
& 85.2$_{\pm0.0}$ & 66.6$_{\pm0.0}$ 
& 85.1$_{\pm0.0}$ & 64.4$_{\pm0.0}$ 
& 82.8$_{\pm0.0}$ & 63.1$_{\pm0.0}$ \\

DFR 
& \underline{92.3}$_{\pm0.2}$ & 88.2$_{\pm0.3}$ 
& 89.3$_{\pm0.2}$ & 87.1$_{\pm1.1}$  
& 86.1$_{\pm0.0}$ & {70.5}$_{\pm0.0}$   
& 85.1$_{\pm0.0}$ & 71.2$_{\pm0.0}$ 
& 83.8$_{\pm0.0}$ & 68.1$_{\pm0.0}$ \\
\midrule

\cellcolor{lightgray}\ladder (\textbf{ours})
& \cellcolor{lightgray}\textbf{93.1}$_{\pm0.8}$ &\cellcolor{lightgray} \textbf{91.4}$_{\pm0.8}$ 
& \cellcolor{lightgray}89.8$_{\pm1.2}$ &\cellcolor{lightgray} \textbf{88.9}$_{\pm0.4}$ 
& \cellcolor{lightgray}\underline{86.8}$_{\pm0.0}$ & \cellcolor{lightgray}\textbf{76.2}$_{\pm0.0}$  
& \cellcolor{lightgray}\underline{85.3}$_{\pm0.0}$ & \cellcolor{lightgray}\textbf{76.4$_{\pm0.0}$}
& \cellcolor{lightgray}\underline{86.2}$_{\pm0.0}$ & \cellcolor{lightgray}\textbf{82.5$_{\pm0.0}$}\\
\bottomrule
\end{tabular}
\end{adjustbox}
\end{table*}

\subsubsection{Extended qualitative results for our slice discovery method on various datasets}
\label{appendix: extended_qual}
\begin{figure*}[t]
\begin{center}
\includegraphics[width=\linewidth]{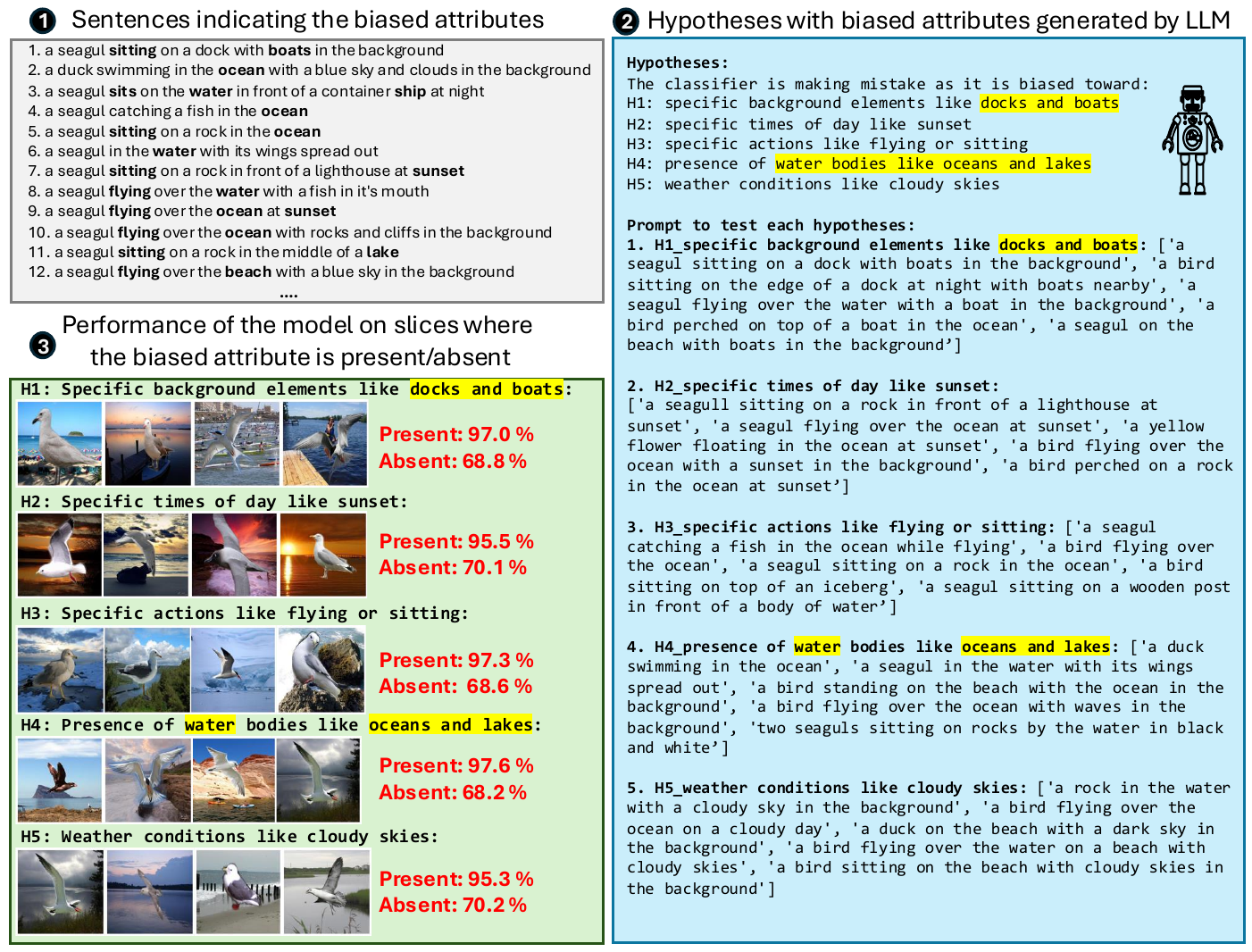}
\caption{
\ladder discovers slices for biased attributes in RN Sup IN1k-based classifier for \textit{waterbird} classification in \textbf{Waterbirds} dataset. This figure details the slice discovery process for biased attributes involving sentence analysis, hypothesis generation by an LLM, and the model's performance on slices where attributes are present or absent, demonstrating how biases affect classifier accuracy. We highlight the hypothesis generated by \ladder that corresponds to the ground truth biased attribute (\eg \texttt{water} for landbirds) in \textbf{yellow}.
}

\label{fig:prompt_waterbirds_cnn}
\end{center}
\end{figure*}

\begin{figure*}[h]
\begin{center}
\includegraphics[width=\linewidth, keepaspectratio]{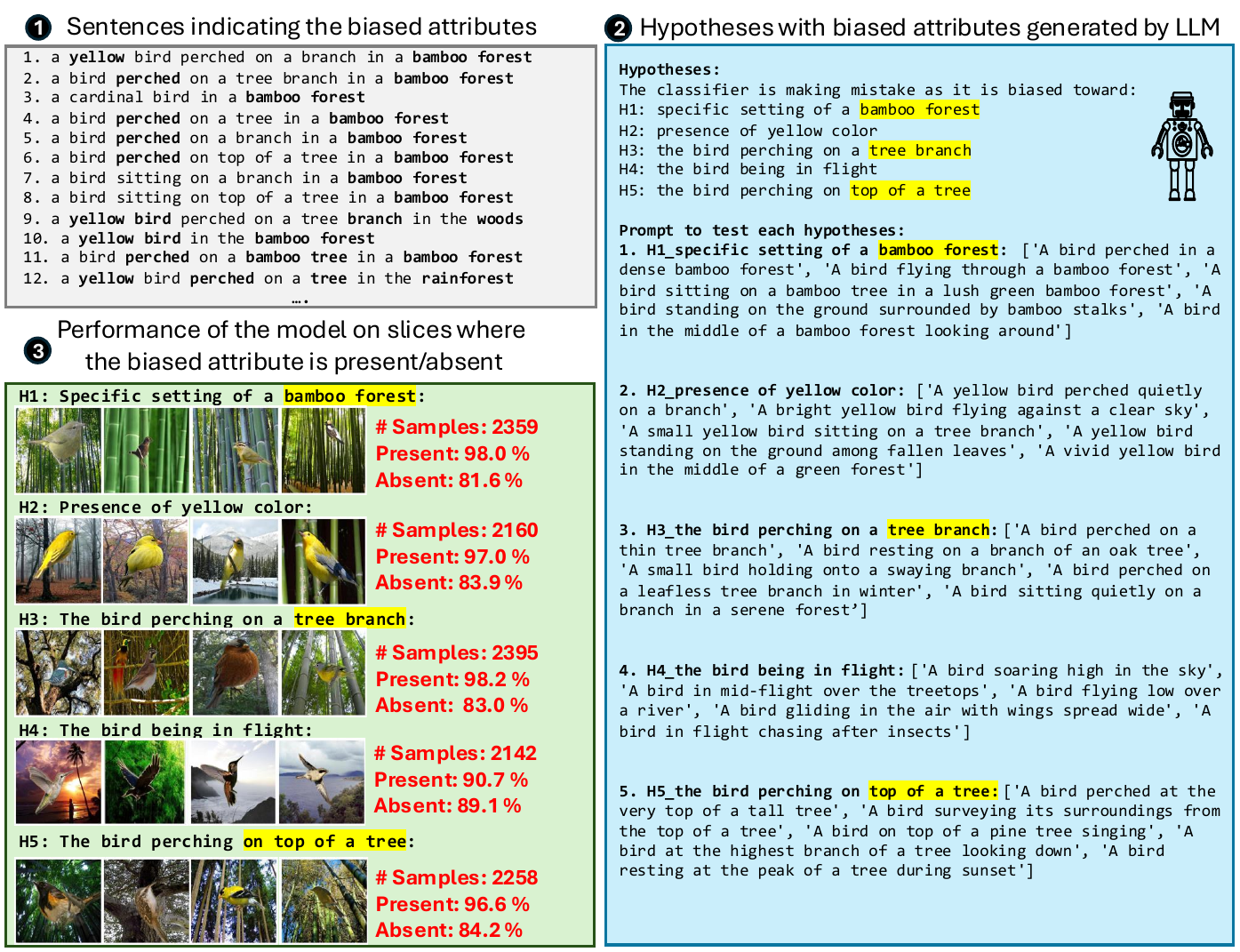} 
\caption{\ladder discovers slices for biased attributes in RN Sup IN1k-based classifier for \textit{landbird} classification in \textbf{Waterbirds} dataset. This figure details the slice discovery process for biased attributes involving sentence analysis, hypothesis generation by an LLM, and the model's performance on slices where attributes are present or absent, demonstrating how biases affect classifier accuracy. We highlight the hypothesis generated by \ladder that corresponds to the ground truth biased attribute (\eg \texttt{land} for landbirds) in \textbf{yellow}.}
\label{fig:qual_landbird_cnn}
\end{center}
\end{figure*}

\begin{figure*}[h]
\begin{center}
\includegraphics[width=\linewidth, keepaspectratio]{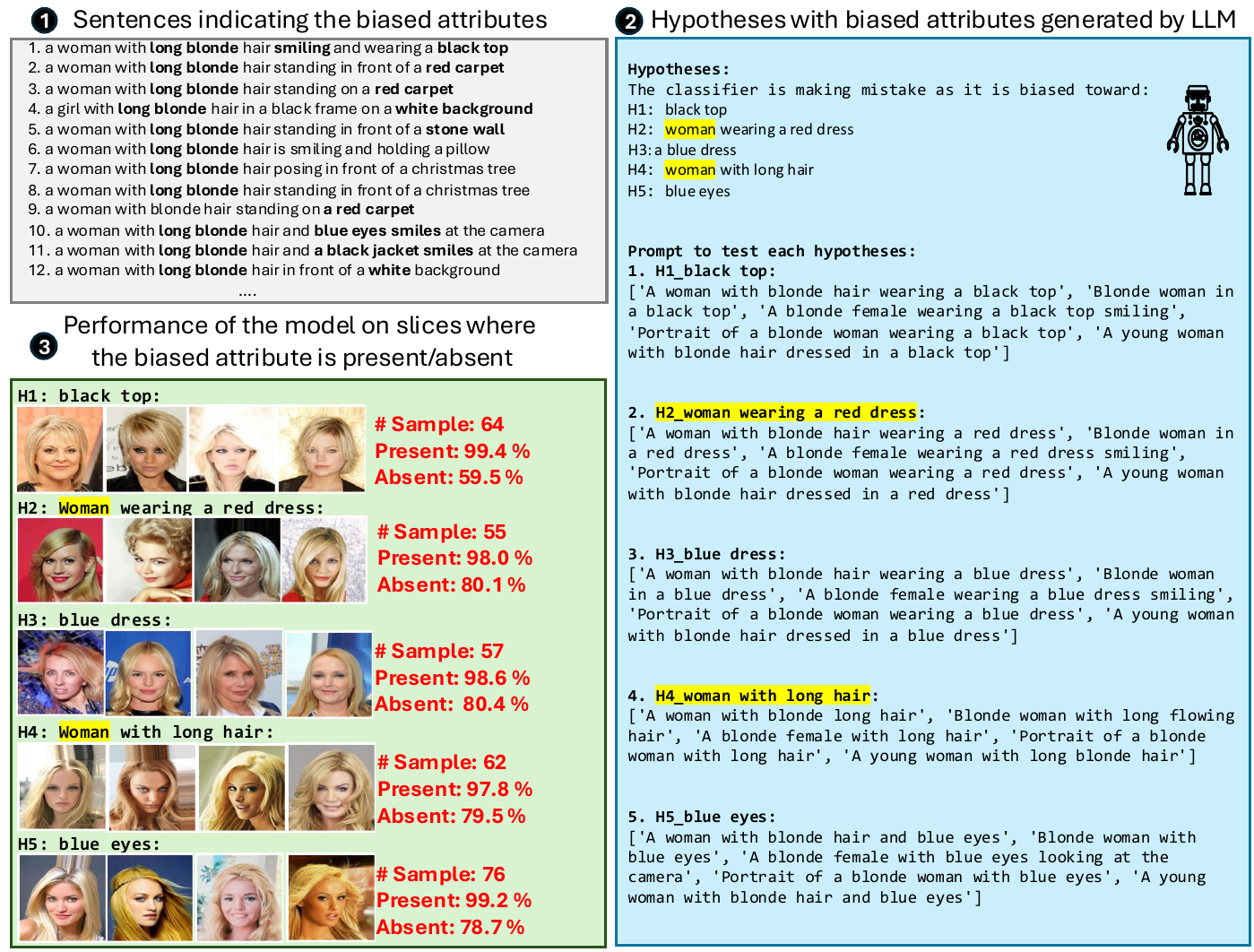} 
\caption{
\ladder discovers slices for biased attributes in RN Sup IN1k-based classifier for \textit{blond} classification in \textbf{CelebA} dataset. This figure details the slice discovery process for biased attributes involving sentence analysis, hypothesis generation by an LLM, and the model's performance on slices where attributes are present or absent, demonstrating how biases affect classifier accuracy. We highlight the hypothesis generated by \ladder that corresponds to the ground truth biased attribute (\eg \texttt{woman} for blond) in \textbf{yellow}.
}
\label{fig:qual_celebA_cnn}
\end{center}
\end{figure*}

\begin{figure*}[h]
\begin{center}
\includegraphics[width=\linewidth, keepaspectratio]
{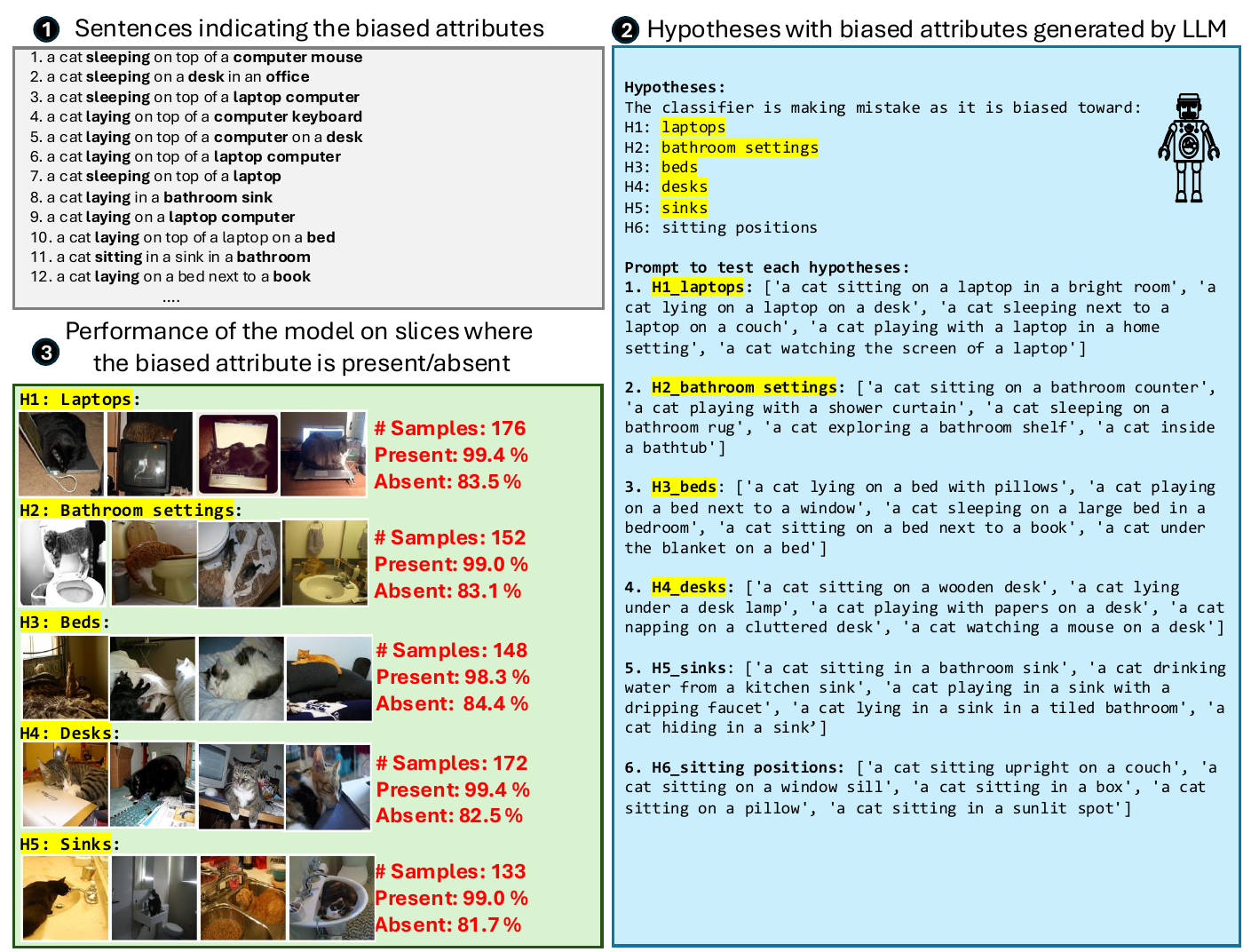}
\caption{
\ladder discovers slices for biased attributes in RN Sup IN1k-based classifier for \textit{cat} classification in \textbf{MetaShift} dataset. This figure details the slice discovery process for biased attributes involving sentence analysis, hypothesis generation by an LLM, and the model's performance on slices where attributes are present or absent, demonstrating how biases affect classifier accuracy. We highlight the hypothesis generated by \ladder that corresponds to the ground truth biased attribute (\eg \texttt{indoor} for cat) in \textbf{yellow}.
}
\label{fig:qual_cat_cnn_ex}
\end{center}
\end{figure*}

\begin{figure*}[h]
\vskip 0.2in
\begin{center}
\includegraphics[width=\linewidth, keepaspectratio]
{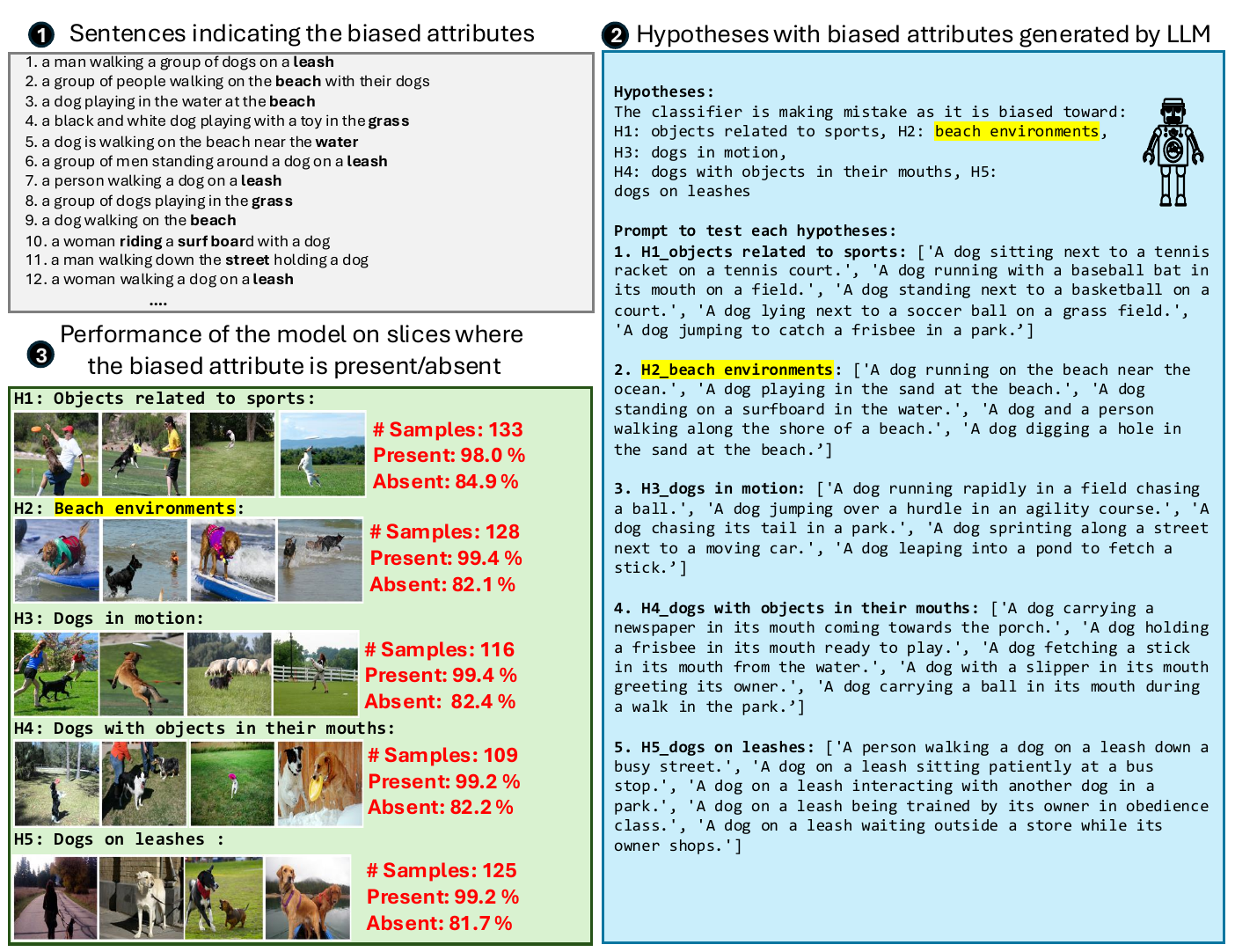}
\caption{
\ladder discovers slices for biased attributes in RN Sup IN1k-based classifier for \textit{dog} classification in \textbf{MetaShift} dataset. This figure details the slice discovery process for biased attributes involving sentence analysis, hypothesis generation by an LLM, and the model's performance on slices where attributes are present or absent, demonstrating how biases affect classifier accuracy. We highlight the hypothesis generated by \ladder that corresponds to the ground truth biased attribute (\eg \texttt{outdoor} for cat) in \textbf{yellow}.}
\label{fig:qual_dog_cnn_ex}
\end{center}
\end{figure*}

\begin{figure*}[t]
\begin{center}
\includegraphics[width=\linewidth]{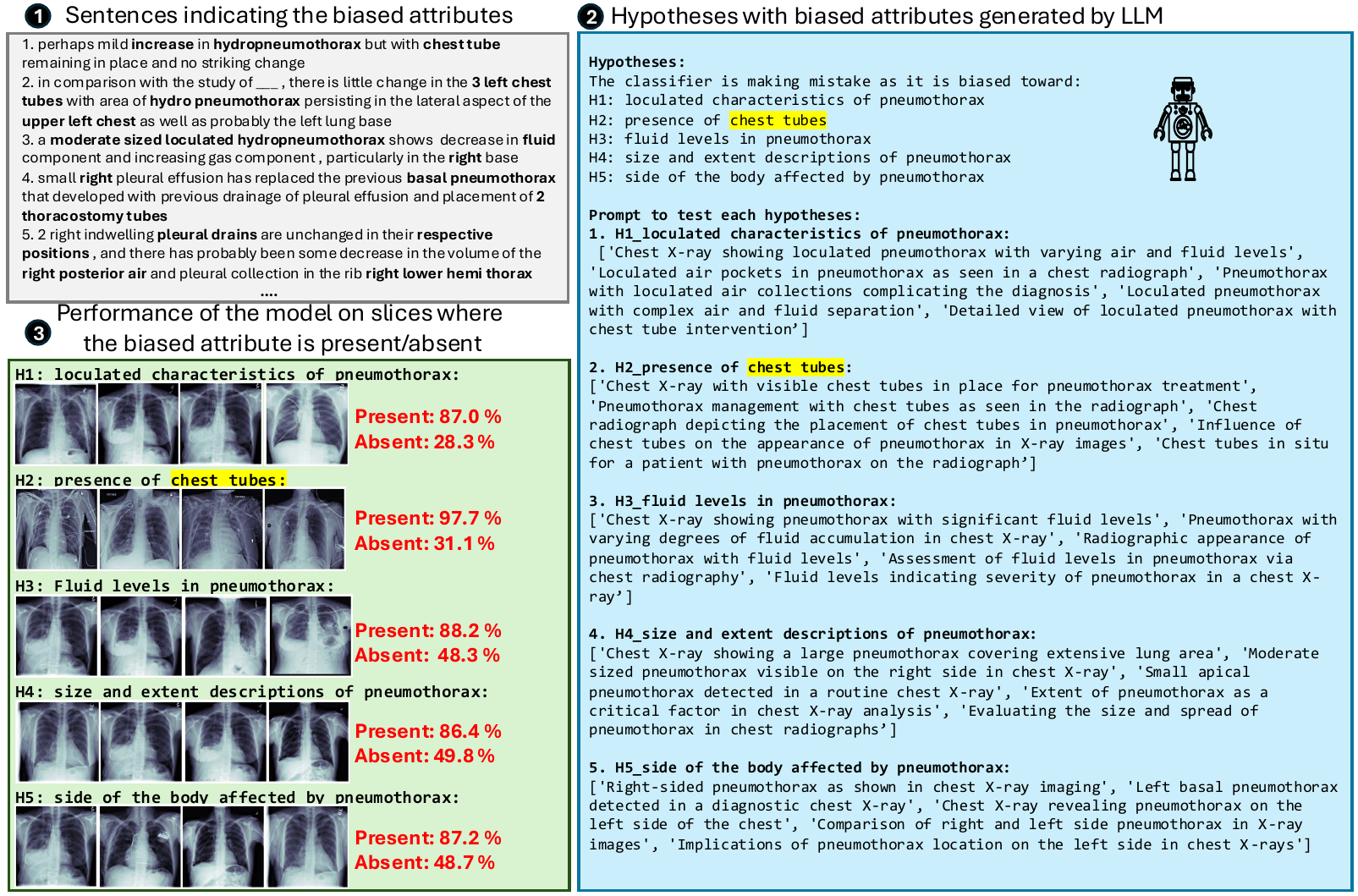}
\caption{
\ladder discovers slices for biased attributes in RN Sup IN1k-based classifier for \textit{pneumothorax} classification in \textbf{NIH-CXR} dataset. This figure details the slice discovery process for biased attributes involving sentence analysis, hypothesis generation by an LLM, and the model's performance on slices where attributes are present or absent, demonstrating how biases affect classifier accuracy. We highlight the hypothesis generated by \ladder that corresponds to the ground truth biased attribute (\eg \texttt{chest-tube} for landbirds) in \textbf{yellow}.
}
\label{fig:prompt_nih_cnn}
\end{center}
\end{figure*}

\begin{figure*}[h]
\vskip 0.2in
\begin{center}
\includegraphics[width=\linewidth, keepaspectratio]{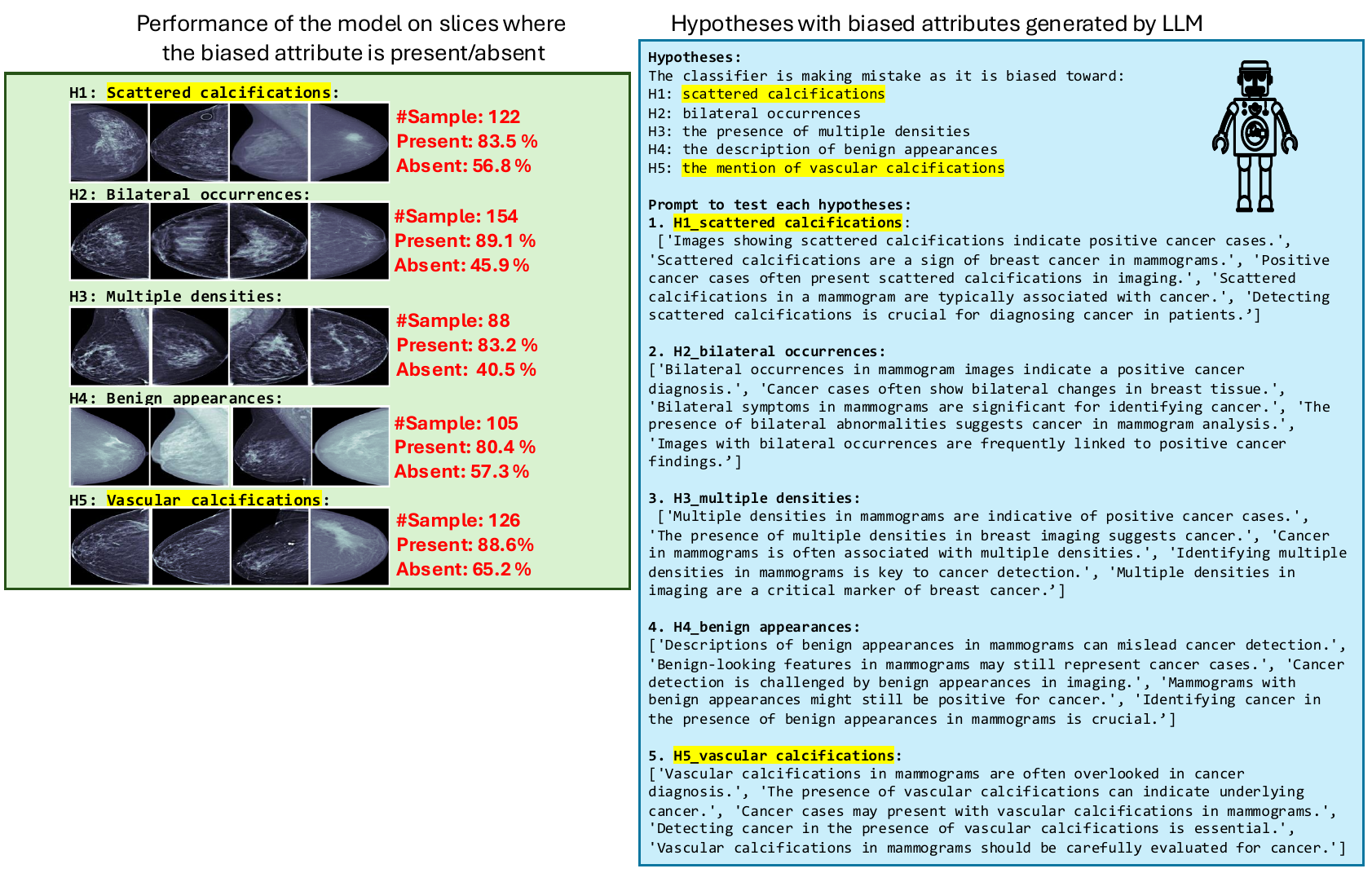}
\caption{
\ladder discovers slices for biased attributes for \textit{cancer} classification in \textbf{RSNA-Mammo} dataset. This figure details the slice discovery process for biased attributes involving sentence analysis, hypothesis generation by an LLM, and the model's performance on slices where attributes are present or absent, demonstrating how biases affect classifier accuracy. We highlight the hypothesis generated by \ladder that corresponds to the ground truth biased attribute (\eg \texttt{calcification} for cancer) in \textbf{yellow}.}
\label{fig:qual_rsna_cnn}
\end{center}
\end{figure*}

Figures~\ref{fig:prompt_nih_cnn} and~\ref{fig:prompt_waterbirds_cnn} report LLM-generated the list of hypotheses and the prompts to test them discussed in the Sec.~\ref{results:nih_waterbirds}.
Figures~\ref{fig:qual_landbird_cnn},~\ref{fig:qual_celebA_cnn},~\ref{fig:qual_cat_cnn_ex},~\ref{fig:qual_dog_cnn_ex}, and~\ref{fig:qual_rsna_cnn} illustrate qualitative results of our method applied on various datasets using RN Sup IN1k models. Specifically, they showcase the classification of pneumothorax patients from NIH, ``landbird'' from the Waterbirds, ``blond'' from CelebA, ``cat'' and ``dog'' from MetaShift, and ``cancer'' from the RSNA-Mammo datasets, respectively. In all the cases, \ladder correctly identifies the hypothesis with true attribute causing biases in the given classifier $f$.

\subsubsection{ Comparing the performance of\ladder for error mitigation across architectures}
\label{app:mitigation_extended}
\begin{table}[ht]
\centering
\caption{Benchmarking error mitigation methods over 3 seeds for ViT models pretrained with IN1k using the supervised method (RN Sup IN1k). We bold-face and underline the best and second-best results, respectively.}
\label{tab:mitigation_VIT}
\resizebox{\columnwidth}{!}{ 
\begin{tabular}{lcccc}
\toprule
\multirow{2.5}{*}{\textbf{Method}} & 
\multicolumn{2}{c}{\textbf{Waterbirds}} 
& \multicolumn{2}{c}{\textbf{CelebA}} \\
\cmidrule(lr){2-3} \cmidrule(lr){4-5}
& \textbf{Mean(\%)} & \textbf{WGA(\%)} 
& \textbf{Mean(\%)} & \textbf{WGA(\%)} \\
\midrule
Vanilla (ERM)
& 82.7$_{\pm1.4}$ & 51.2$_{\pm1.3}$ 
& 95.2$_{\pm0.4}$ & 46.8$_{\pm1.1}$ \\
\midrule

Mixup
& 81.8$_{\pm0.4}$ & 44.9$_{\pm0.3}$
& \textbf{95.8}$_{\pm0.3}$ & 48.3$_{\pm0.3}$ \\
IRM
& 79.8$_{\pm0.3}$ & 54.5$_{\pm0.3}$  
& 85.1$_{\pm1.2}$ & 48.7$_{\pm0.3}$ \\
MMD
& 83.6$_{\pm2.7}$ & 42.5$_{\pm1.1}$ 
& 95.6$_{\pm0.4}$ & 54.2$_{\pm0.4}$ \\
\midrule
JTT
& 81.7$_{\pm0.5}$ & 49.1$_{\pm0.5}$
& 94.8$_{\pm0.3}$ & 52.7$_{\pm0.6}$ \\
GroupDRO
& 82.2$_{\pm0.8}$ & 53.1 $_{\pm1.2}$
& 93.5$_{\pm0.1}$ & 80.1$_{\pm0.4}$ \\
CVaRDRO
& 83.5$_{\pm0.3}$ & 46.6$_{\pm2.8}$
& 95.6$_{\pm0.1}$ & 55.1$_{\pm1.8}$ \\
LISA
& 83.7$_{\pm0.1}$ & 48.8$_{\pm0.1}$ 
& 95.6$_{\pm0.2}$ & 60.2$_{\pm0.1}$ \\
DFR
& 85.0$_{\pm0.3}$ & 76.2$_{\pm0.3}$
& 91.3$_{\pm1.1}$ & 81.1$_{\pm0.5}$ \\
\midrule
\cellcolor{lightgray}\ladder (\textbf{ours})
& \cellcolor{lightgray}\textbf{85.3}$_{\pm0.5}$ & \cellcolor{lightgray}\textbf{84.5$_{\pm0.4}$}
& \cellcolor{lightgray}90.7$_{\pm0.1}$ &\cellcolor{lightgray}83.4$_{\pm0.1}$ \\
\bottomrule
\end{tabular}
}
\end{table}

\begin{figure*}[h]
\begin{center}
\includegraphics{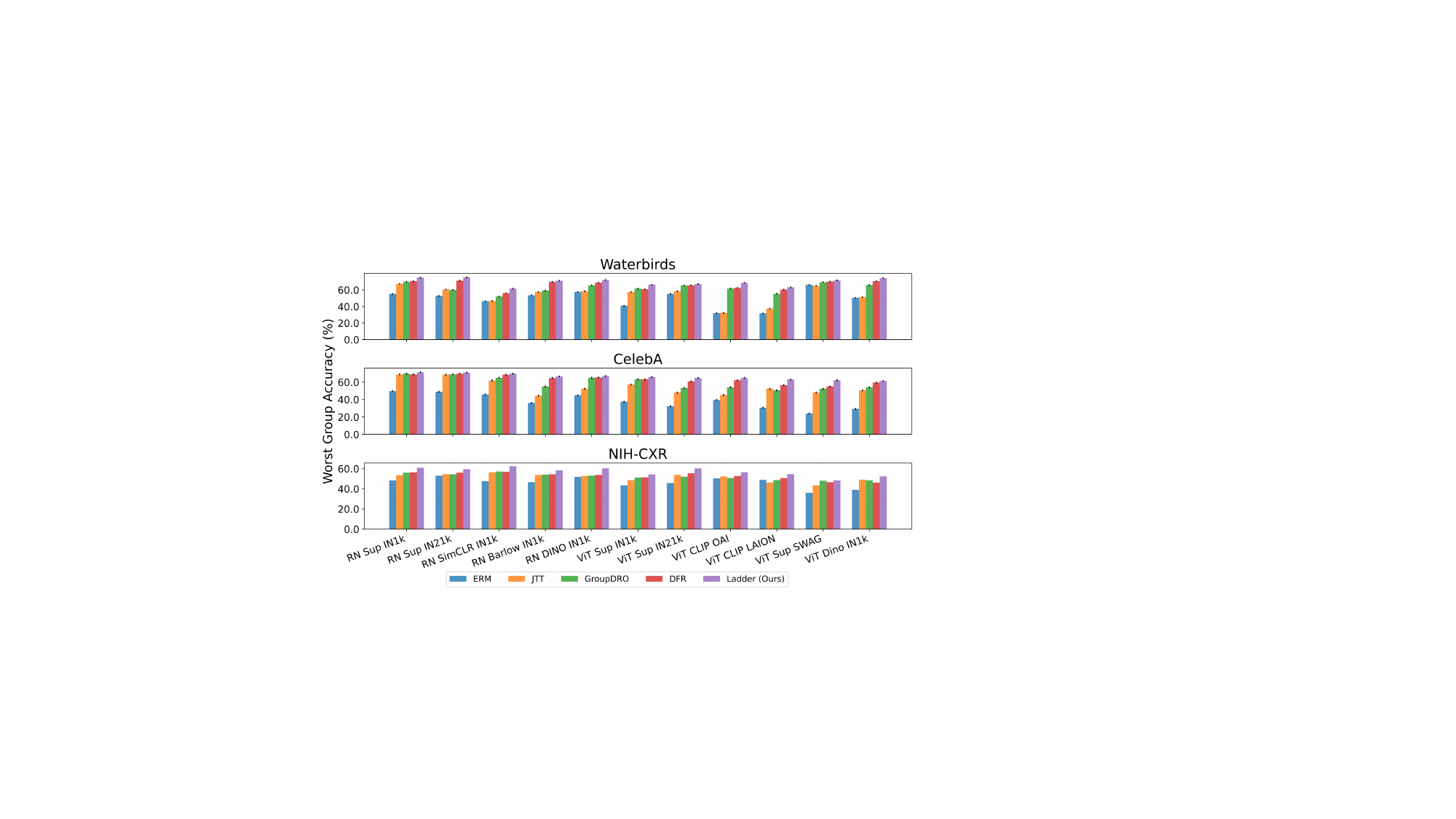}
\caption{WGA across bias mitigation methods.\ladder consistently outperforms other bias mitigation baselines (ERM, JTT, GroupDRO, and DFR) across different model architectures and pre-training strategies.} 
\label{fig:wga_all}
\end{center}
\end{figure*}

Tab.~\ref{tab:app_mitigation_RN50} compares~\ladder with additional bias mitigation baselines for CNN-based models. Tab.~\ref{tab:mitigation_VIT} compares different error mitigation algorithms for ViT Sup IN1K-based models ($f$), for all the SOTA mitigation baselines discussed in Appendix~\ref{appendix_rr_mitigation}. For natural images (Waterbirds and CelebA), we report mean accuracy. For medical images (NIH,
RSNA and VinDr), we report mean AUROC. Fig.~\ref{fig:wga_all} reports the WGA and shows that\ladder outperforms the other slice discovery baselines across the different architectures and pre-training strategies.



\subsubsection{ Application: Improvement on the zero-shot accuracy of Vision Language models using the attributes from the extracted hypothesis by\ladder}
\label{app: zs_boost}
To evaluate the impact of LADDER's attribute-based slice discovery on zero-shot performance, we conducted experiments using a CLIP-based vision-language model across multiple datasets. LADDER extracts fine-grained attributes from error-prone data slices, which we incorporated as detailed prompts for zero-shot classification. These prompts were generated from hypotheses produced by the LADDER framework and reflect nuanced characteristics of the data that a model might otherwise overlook. We compare these attribute-driven prompts against standard, baseline prompts typically used for zero-shot tasks.

\textbf{Experimental Process.}
For each dataset, we implemented two types of zero-shot prompts:
\begin{itemize}
    \item \textbf{Baseline prompts}: CLIP-based prompts~\citep{radford2021learning} \eg [\texttt{a photo of a landbird} and \texttt{a photo of a waterbird}] for the Waterbirds dataset for natural images, CXR-CLIP~\citep{you2023cxr} prompts \eg [\texttt{no pneumothorax}, \texttt{pneumothorax}] for NIH, Mammo-CLIP~\citep{ghosh2024mammo} prompts \eg [\texttt{\{no cancer, no malignancy\}}, \texttt{\{cancer, malignancy\}}] for RSNA-Mammo and VinDr-Mammo.
    \item \textbf{LADDER-derived prompts}: These prompts were generated based on the attributes extracted from LADDER’s hypotheses, providing a more detailed description of the data. For example, in the Waterbirds dataset, we used prompts like \texttt{a photo of a waterbird on docks and boats} or \texttt{a photo of a landbird inside on bamboo forest}. In this experiment, we use the attributes from the hypotheses extracted from RN Sup IN1k (Resnet 50 pretrained with ImageNet 1K and supervised learning) classifier.
\end{itemize}

We evaluated the zero-shot classification performance of the model using both prompt types. The results are shown in Tab.~\ref{tab:zeroshot_accuracy}.

\textbf{Results.} The results demonstrate a significant improvement in zero-shot accuracy when using LADDER-extracted attributes as prompts. Across all datasets, the attribute-driven prompts outperformed the baseline, indicating the effectiveness of using detailed, hypothesis-driven attributes to enhance zero-shot performance. In the \textbf{Waterbirds} dataset, LADDER prompts improved accuracy by +8.56\%, rising from 50.40\% with baseline prompts to 58.96\% with LADDER attributes. The improvement was even more pronounced for the \textbf{NIH} dataset, with a +19.05\% gain (49.17\% to 68.22\%). The \textbf{RSNA} dataset also saw a notable improvement, with a +5.81\% gain in accuracy (60.17\% to 65.98\%). The improvements for \textbf{CelebA} (+0.32\%) and \textbf{VinDr} (+1.41\%) were more modest but still indicate that using LADDER’s attribute-based prompts provides consistent gains across various domains. These results highlight the ability of LADDER to extract meaningful attributes that guide the vision-language model to more accurate predictions, even in zero-shot settings where explicit training on the target data is absent. By leveraging these hypotheses, LADDER enables more precise alignment between image representations and class descriptions, significantly enhancing zero-shot performance.

\label{zs_clip}
\begin{table}[h]
\centering
\scriptsize
\caption{Application: Boost in Zero-shot accuracy results using attributes from the hypotheses extracted from RN Sup IN1k (Resnet 50 pretrained with ImageNet 1K and supervised learning) classifier}
\begin{tabular}{lccc}
\toprule
Dataset & CLIP Prompts & \ladder Hypotheses & Gain \\
\midrule
Waterbirds & 50.40 & \textbf{58.96} & +8.56 $\uparrow$ \\
CelebA & 86.69 &  \textbf{87.01} & +0.32 $\uparrow$ \\
NIH & 49.17 &  \textbf{68.22} & +19.05 $\uparrow$ \\
RSNA & 60.17 &  \textbf{65.98} & +5.81 $\uparrow$ \\
VinDr & 90.92 &  \textbf{92.33} & +1.41 $\uparrow$ \\
\bottomrule
\end{tabular}
\label{tab:zeroshot_accuracy}
\end{table}

\subsubsection{ CLIP score comparison of various attributes extracted by\ladder}
\label{clip_score}
Refer to Fig.~\ref{fig:clip_score} for the CLIP scores (discussed in Appendix~\ref{appendix: clip_score_def}) of various attributes extracted from the hypotheses by\ladder. For \eg the correctly classified samples for the waterbird class in the Waterbirds dataset have a bias on the water-related backgrounds. As a result, the CLIP score of \texttt{ocean, boat, lake} is high. We observe consistent results for other datasets as well.

\begin{figure*}[h]
\begin{center}
\includegraphics[width=\linewidth]{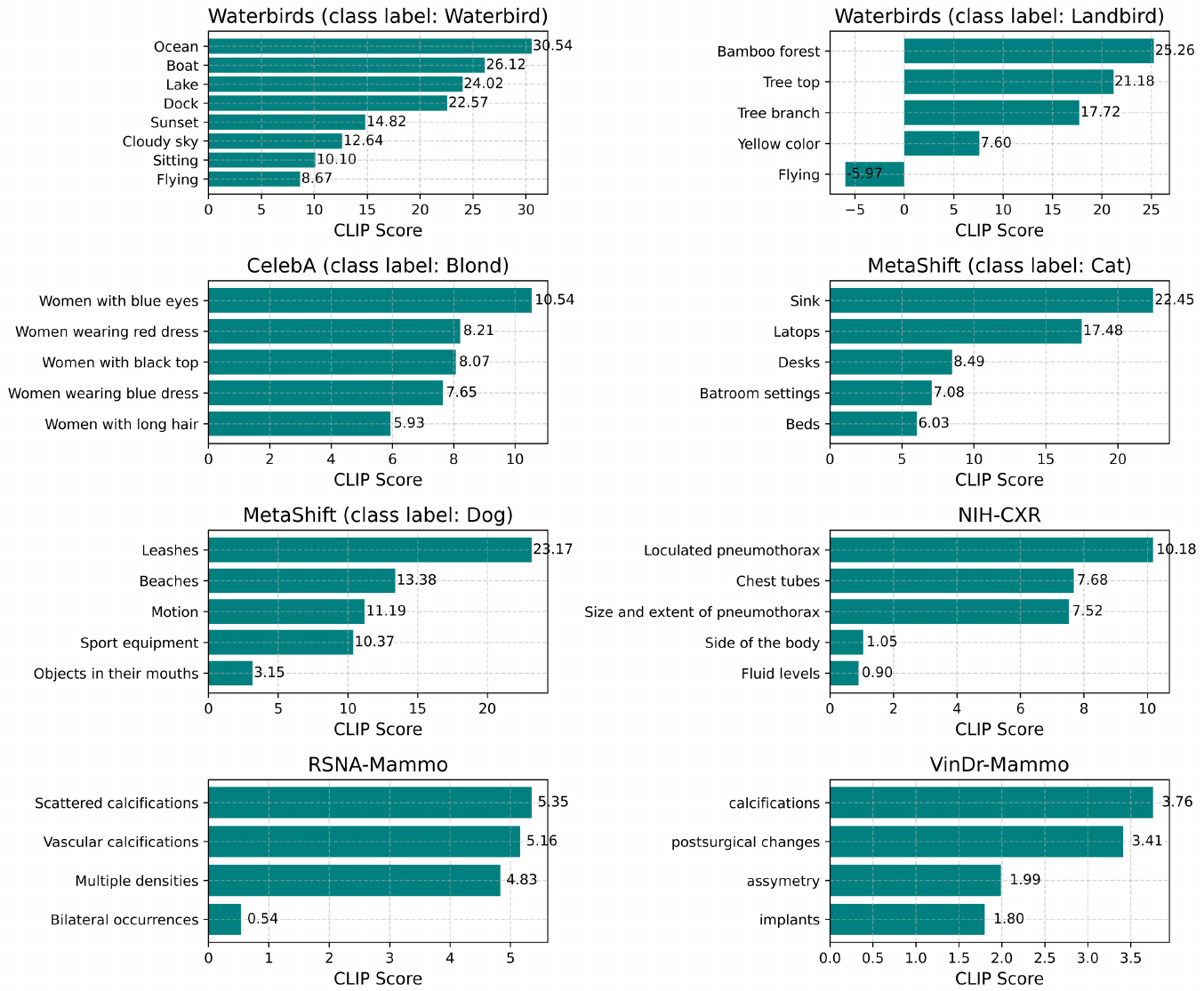}
\caption{CLIP Score(Appendix~\ref{appendix: clip_score_def}) for various attributes extracted from the hypotheses by\ladder. CLIP scores of the attributes are high signifying that they induce biases on the correctly classified samples.} 
\label{fig:clip_score}
\end{center}
\end{figure*}




\subsubsection{ Improvement on different slices of UrbanCars benchmark}
\label{app: whac}
Tab.~\ref{tab:whacamole} shows that\ladder achieves higher accuracy compared to the Whac-A-Mole method\citep{li2023whac} across multiple shortcut benchmarks on the Urbancars dataset, without prior knowledge of the number or types of possible shortcuts.

\begin{table}[h]
\centering
\scriptsize
\setlength{\tabcolsep}{2pt} 
\caption{\ladder achieves higher accuracy compared to the Whac-A-Mole method~\citep{li2023whac} across multiple shortcut benchmarks on the Urbancars dataset without prior knowledge of the number or types of possible shortcuts.}
\begin{tabular}{lcccc}
\toprule
Method & Mean Acc & BG gap & CoObj Gap & BG+CoObj Gap \\
\midrule
ERM & 96.4 & -15.3 & -11.2 & -69.2 \\
Whac-A-Mole & 95.2 & -2.4 & -2.9 & -5.8 \\
\ladder & 92.2 & -1.1 & -1.6 & -3.8 \\
\bottomrule
\end{tabular}
\label{tab:whacamole}
\end{table}

\subsubsection{ Extended results on discovered hypothesis by\ladder for various architectures and pre-training methods}
\label{appendix: slices_arch_pretraining}
\begin{figure*}[h]
\begin{center}
\includegraphics[width=\linewidth]{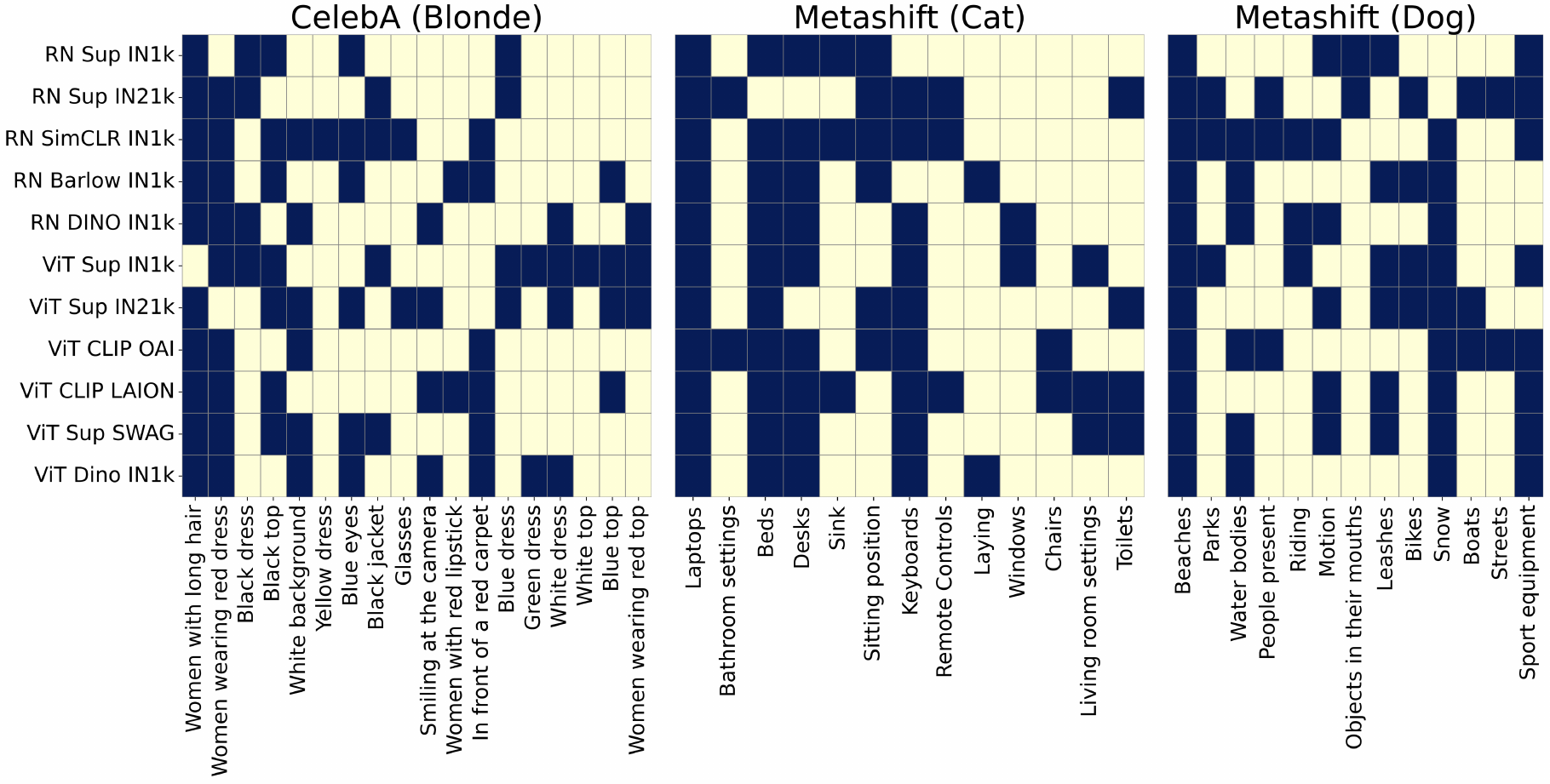}
\caption{\ladder accurately captures various sources of bias, regardless of the underlying architectures or pre-training methods for the CelebA and Metashift datasets. Bright colors indicate attributes in \ladder's hypotheses, while light colors indicate their absence.} 
\label{fig:hyp_supp_extended}
\end{center}
\end{figure*}

Fig.~\ref{fig:hyp_supp_extended} illustrates additional results for the CelebA and Metashift datasets, demonstrating that\ladder accurately captures various sources of bias, regardless of the underlying architectures or pre-training methods.

\subsubsection{ Results on Imagenet}
\label{sec: imagenet}
Tables~\ref{tab: imagenet_stetho},~\ref{tab: imagenet_ant},~\ref{tab: imagenet_horizontal_bar}
shows that\ladder identifies unique biases for the Imagenet dataset for a stethoscope, ant, and horizontal bar, respectively.


\subsubsection{ Ablation 1: WGA of\ladder using other captioning methods}
\label{WGA_captioning}
Tab.~\ref{tab:captioning_ablation} presents an ablation study evaluating the effect of various captioning models on\ladder's performance in mitigating biases. The quality of captions directly affects \ladder's ability to effectively generate hypotheses, as these captions are analyzed by LLMs to identify biased attributes contributing to model errors.\ladder then pseudo-labels these attributes to systematically mitigate the identified biases. We consider different captioning models, including BLIP~\citep{li2022blip}, BLIP2~\citep{li2023blip}, ClipCap~\citep{mokady2021clipcap}, and GPT-4o~\citep{wu2024gpt}, with \textbf{ResNet Sup IN1k} as the classifier.

The results indicate that the more advanced captioning model, GPT-4o, significantly improves \ladder's performance, achieving the highest Worst Group Accuracy (WGA) and mean accuracy across both datasets. Specifically, GPT-4o achieves a WGA of 94.5\% on Waterbirds and 91.9\% on CelebA, which is substantially better than the other models. BLIP and BLIP2 demonstrate comparable results, with BLIP slightly outperforming BLIP2 in the Waterbirds dataset, while BLIP2 performs better on CelebA in WGA. In contrast, ClipCap consistently yields the lowest scores, implying that simpler captioning methods are less effective for enhancing \ladder's bias identification capabilities.
Overall, the results underscore the importance of selecting a high-quality captioning model to maximize \ladder's effectiveness. While more sophisticated models like GPT-4o entail higher costs, their significant impact on bias mitigation performance, particularly on WGA, makes them an indispensable choice in scenarios where accuracy is critical.


\subsubsection{   Ablation 2: Slice discovery by\ladder using different LLMs}
\begin{figure*}[h]
\begin{center}
\includegraphics[height=1.1\linewidth]{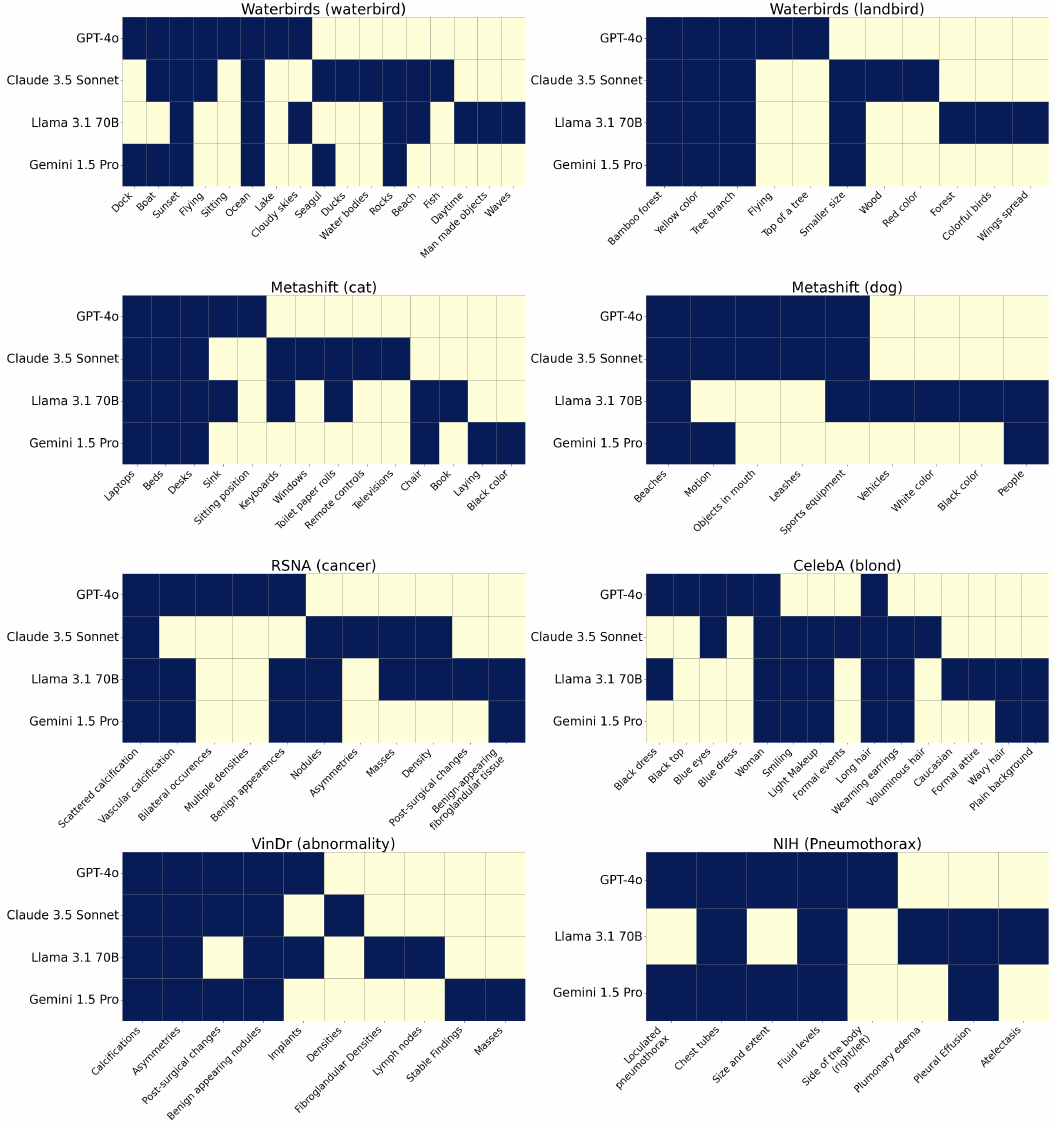}
\caption{Ablation 2: Attributes identified by different LLMs while generating hypotheses across datasets for bias identification: RN Sup IN1k for natural images and CXRs, and EN-B5 for mammograms. Each LLM (GPT-4o, Claude 3.5 Sonnet, LlaMA 3.1 70B, Gemini 1.5 Pro) focuses on distinct attributes, yet the overall hypotheses are consistent across datasets, showing\ladder's robust bias detection. Bright colors indicate attributes in \ladder's hypotheses, while light colors indicate their absence. Following MIMIC's regulations, we use Gemini 1.5 Pro (via Vertex AI on Google Cloud Platform), GPT-4o via Azure OpenAI service, and Llama 3.1 70B (running locally) for NIH. Bright colors indicate attributes in\ladder's hypotheses.} 
\label{fig:LLM_slice}
\end{center}
\end{figure*}

In this ablation study, we explore how different LLMs impact the effectiveness of\ladder in discovering data slices and generating hypotheses for bias identification. We aim to discover the biases from RN Sup IN1k classifier for natural images and CXRs, and EN-B5 classifier for mammograms. We utilize four LLMs: GPT-4o, Claude 3.5 Sonnet, LLaMA 3.1 70B, and Gemini 1.5 Pro. Fig.~\ref{fig:LLM_slice} illustrates the different attributes these models highlight across multiple datasets, including Waterbirds, CelebA, NIH, RSNA, VinDr, and MetaShift.
Each LLM aims to extract a hypothesis related to an attribute, signifying the classifier's mistake. These attributes potentially lead to systematic model biases. As shown in Fig.~\ref{fig:LLM_slice}, each LLM focuses on distinct subsets of attributes, reflecting their unique interpretation capabilities. Despite these differences, there is significant overlap in the overall hypotheses generated across the models, indicating consistency in identifying the attributes contributing to model errors.

For instance, in the Waterbirds dataset, all LLMs frequently highlight attributes like \texttt{ocean} and \texttt{boat} for the waterbird class and \texttt{bamboo forest} and \texttt{tree branch} for the landbird class. These attributes align closely with the ground truth bias in this dataset, which relates to water and land backgrounds being associated with the respective bird classes. This suggests that LLMs effectively identify these underlying environmental biases that lead to systematic errors.
Similarly, in medical datasets, such as NIH-CXR for pneumothorax, all LLMs consistently highlight \texttt{chest tube} as a common attribute for misclassified samples. This reflects a true bias, as the presence of a chest tube often strongly correlates with pneumothorax cases. Identifying this attribute helps understand the systematic bias that models may develop when chest tubes are spuriously correlated in pneumothorax images.

This consistency across various LLMs demonstrates the robustness of\ladder for systematic bias detection, irrespective of the underlying LLM used. The results highlight that\ladder is effective at leveraging the strengths of different LLMs to produce meaningful insights into model behavior, regardless of which LLM is utilized. Moreover, it emphasizes the versatility of using LLMs for extracting domain-specific attributes—whether the focus is on natural images, chest X-rays, or mammography scans -- while maintaining cost efficiency and avoiding manual annotation.
Overall, this ablation shows that the specific choice of LLM slightly influences which attributes are emphasized, but all models effectively support the generation of comprehensive hypotheses that capture the biases inherent in different datasets.
\label{App:slice_LLM}

\subsubsection{   Ablation 3: WGA by\ladder using the hypothesis by different LLMs}
Fig.~\ref{fig:LLM_WGA} illustrates the worst group accuracy (WGA) achieved across multiple datasets when utilizing\ladder to mitigate biases with different LLMs. The LLMs compared in this study include Claude 3.5 Sonnet, LLaMA 3.1 70B, Gemini 1.5 Pro, and GPT-4o. We consider the RN Sup IN1k classifier for natural images and CXRs, as well as the EN-B5 classifier for mammograms.
The primary aim of this ablation is to assess how well\ladder can mitigate biases when generating hypotheses using different LLMs. 
As shown in Fig.~\ref{fig:LLM_WGA}, the WGA values remain consistently high across all LLMs, indicating that\ladder is effective in mitigating biases irrespective of the choice of LLM for hypothesis generation. Specifically, all LLMs achieve WGA scores of over 80\% for most datasets, with only slight variations between models. This consistency demonstrates the robustness of\ladder in leveraging different LLMs to address model biases effectively.
For datasets like Waterbirds and CelebA, the performance across all LLMs is nearly identical, suggesting that the generated hypotheses successfully capture the underlying biases and lead to similar improvements in fairness. In medical datasets, such as NIH and RSNA, the trend is also maintained, with LLMs like GPT-4o and Gemini 1.5 Pro achieving better results than other LLMs.
These findings emphasize that the specific choice of LLM has only a minor impact on the overall ability of \ladder to mitigate bias. This makes \ladder a flexible and cost-effective solution, as it can work effectively with a range of LLMs, each with different computational costs and capabilities. Using different LLMs ensures flexibility based on resource availability while effectively identifying and mitigating dataset biases.
\label{WGA_LLM}


\subsubsection{ Ablation 4: Overall cost and choice of LLMs}
\label{sec:cost_LLM}
Tab.~\ref{tab:llm_token_breakdown} shows the cost of using various LLMs. Each row shows the total breakdown for an LLM extracting hypotheses across all 6 datasets, using RN Sup IN1k (natural images or CXRs) and EN-B5 (mammograms).\ladder invokes LLM once using sentences only (no images). The total cost incurred is $\sim$\$28 across all architectures and pretraining used in the experiments. Thus, LLMs are far more cost-effective than developing new tagging models for unexplored domains \eg radiology, or manually annotating shortcuts.
Fig.~\ref{fig:LLM_slice} in Appendix~\ref{App:slice_LLM} shows the attributes identified by each LLM while generating hypotheses. Different LLMs capture distinct sets of attributes, yet substantial overlap exists, with many attributes consistently revealing actual biases across models. Ablation studies in Appendix~\ref{WGA_LLM} indicate that using different LLMs to compute WGA shows that Gemini and GPT-4o achieve higher WGA for medical images than the others.

\subsubsection{ Ablation 5: Choice of VLR on\ladder}
\label{sec:ablation_cxr}
Fig.\ref{fig:cxr_ablations} demonstrates that{\ladder} consistently detects well-known biases in CXRs, such as \texttt{chest tube}, across various VLRs (CXR-CLIP (SwinT), GLORIA~\cite{huang2021gloria}, MedCLIP (SwinT)~\cite{wang2022medclip}, and MedKLIP (SwinT)~\cite{wu2023medklip}) on the NIH dataset. This consistency suggests that the choice of VLR does not significantly impact{\ladder}'s ability to identify biased attributes.

\begin{table}[h]
\centering
\scriptsize
\setlength{\tabcolsep}{2pt} 
\renewcommand{\arraystretch}{1.1} 
\caption{\ladder identifies unique biases in \textbf{ImageNet} for the ``Stethoscope'' class. The table shows accuracy for subpopulations where the hypothesis failed (Error Slice) and where it passed (Bias-Aligned).}
\begin{tabular}{p{3cm}>{\centering\arraybackslash}p{2cm}>{\centering\arraybackslash}p{2cm}}

\toprule
\textbf{Biases} & 
\parbox[c]{2cm}{\centering \textbf{Accuracy of the subpopulation} \\ \textbf{where hypothesis failed} \\ \textbf{(Error Slice) \\(\%)}} & 
\parbox[c]{2cm}{\centering \textbf{Accuracy of the subpopulation} \\ \textbf{where hypothesis passed} \\ \textbf{(Bias-Aligned) \\(\%)}} \\ 
\midrule
Littmann branding                     & 51.3 & 95.2 \\
Dual-head stethoscopes                & 53.7 & 95.2 \\
Medical settings          & 51.3 & 93.3 \\
Colors \eg yellow or copper & 55.6 & 87.8 \\
Children interacting with stethoscopes & 58.2 & 93.6 \\
\bottomrule
\label{tab: imagenet_stetho}
 \vspace{-2em}
\end{tabular}
\end{table}

\begin{table}[h]
\centering
\scriptsize
\setlength{\tabcolsep}{2pt} 
\renewcommand{\arraystretch}{1.1} 
\caption{\ladder identifies unique biases in \textbf{ImageNet} for the ``Ant'' class. The table shows accuracy for subpopulations where the hypothesis failed (Error Slice) and where it passed (Bias-Aligned).}
\begin{tabular}{p{2.3cm}>{\centering\arraybackslash}p{2cm}>{\centering\arraybackslash}p{2cm}}
\toprule
\textbf{Biases} & 
\parbox[c]{2cm}{\centering \textbf{Accuracy of the subpopulation} \\ \textbf{where hypothesis failed} \\ \textbf{(Error Slice) \\(\%)}} & 
\parbox[c]{2cm}{\centering \textbf{Accuracy of the subpopulation} \\ \textbf{where hypothesis passed} \\ \textbf{(Bias-Aligned) \\(\%)}} \\ 
\midrule
Close up settings                    & 62.6 & 73.3 \\
Textured surface                & 59.6 & 74.5 \\
Green Leaves          & 67.5 & 76.5 \\
Yellow flower & 62.4 & 69.8 \\
Black ant & 63.8 & 73.1 \\
\bottomrule
\label{tab: imagenet_ant}
 \vspace{-2.5em}
\end{tabular}
\end{table}

\begin{table}[h]
\centering
\scriptsize
\setlength{\tabcolsep}{2pt} 
\vspace{-2.6em}
\renewcommand{\arraystretch}{1.1} 
\caption{\ladder identifies unique biases in \textbf{ImageNet} for the ``Horizontal bar'' class. The table shows accuracy for subpopulations where the hypothesis failed (Error Slice) and where it passed (Bias-Aligned).}
\begin{tabular}{p{2.3cm}>{\centering\arraybackslash}p{2cm}>{\centering\arraybackslash}p{2cm}}
\toprule
\textbf{Biases} & 
\parbox[c]{2cm}{\centering \textbf{Accuracy of the subpopulation} \\ \textbf{where hypothesis failed} \\ \textbf{(Error Slice) \\(\%)}} & 
\parbox[c]{2cm}{\centering \textbf{Accuracy of the subpopulation} \\ \textbf{where hypothesis passed} \\ \textbf{(Bias-Aligned) \\(\%)}} \\ 
\midrule
Child                   & 66.4 & 82.4 \\
Playground                & 61.4 & 82.7 \\
Green Leaves          & 67.7 & 76.5 \\
Yellow flower & 62.5 & 69.8 \\
Black ant & 63.5 & 73.8 \\
\bottomrule
\label{tab: imagenet_horizontal_bar}

\end{tabular}
\end{table}

\end{document}